%% file: AGMixup_aaai25_arxiv.tex
\pdfoutput=1%For ARXIV
\documentclass[letterpaper]{article} % DO NOT CHANGE THIS
\usepackage{aaai25}  % DO NOT CHANGE THIS
\usepackage{times}  % DO NOT CHANGE THIS
\usepackage{helvet}  % DO NOT CHANGE THIS
\usepackage{courier}  % DO NOT CHANGE THIS
\usepackage[hyphens]{url}  % DO NOT CHANGE THIS
\usepackage{graphicx} % DO NOT CHANGE THIS
\usepackage{natbib}  % DO NOT CHANGE THIS AND DO NOT ADD ANY OPTIONS TO IT
\usepackage{caption} % DO NOT CHANGE THIS AND DO NOT ADD ANY OPTIONS TO IT
\usepackage{algorithm}
\usepackage{algorithmic}

\usepackage{newfloat}
\usepackage{listings}
\DeclareCaptionStyle{ruled}{labelfont=normalfont,labelsep=colon,strut=off} % DO NOT CHANGE THIS
\floatstyle{ruled}
\newfloat{listing}{tb}{lst}{}
\floatname{listing}{Listing}
\usepackage{xspace}
\usepackage{amsmath}
\usepackage{amssymb}
\usepackage{amsthm}
\usepackage{booktabs}
\usepackage{caption}
\usepackage{subfigure}
\usepackage{graphicx}
\usepackage{colortbl}
\usepackage{multirow}
\usepackage{multicol}
\usepackage{listings}
\usepackage[table,xcdraw]{xcolor}  
\usepackage{arydshln}% 表格虚线
\newcommand\agmixup{\texttt{AGMixup}\@\xspace}
\newcommand{\aVec}[1]{\mathbf{#1}}
\newcommand{\data}{\mathcal{D}}

\newcommand{\mixpoint}[2]{\tilde{\aVec{#1}}_{#2}}
\newcommand{\labelset}{\mathcal{D}_{L}}

\DeclareMathOperator{\betadist}{Beta}

\newcommand{\imp}[1]{#1\% $\uparrow$}
\newcommand{\dec}[1]{#1\% $\downarrow$}
\newcommand{\zero}{0.00\% $\uparrow$}

\title{AGMixup: Adaptive Graph Mixup for Semi-supervised Node Classification}
\author{
    Weigang Lu\textsuperscript{\rm 1},
    Ziyu Guan\textsuperscript{\rm 1}\thanks{Corresponding Author},
    Wei Zhao\textsuperscript{\rm 1},
    Yaming Yang\textsuperscript{\rm 1},
    Yibing Zhan\textsuperscript{\rm 2},
    Yiheng Lu\textsuperscript{\rm 1},
    Dapeng Tao\textsuperscript{\rm 3}
}
\affiliations{
    \textsuperscript{\rm 1}School of Computer Science and Technology, Xidian University, Xi'an, China\\
    \textsuperscript{\rm 2}JD Explore Academy, Xidian University, Beijing, China\\
    \textsuperscript{\rm 3}School of Information Science and Engineering, Yunnan University, Kunming, China
    \{wglu@.stu., zyguan@, ywzhao@mail., yym@,\}xidan.edu.cn, zhanyibing@jd.com, dapeng.tao@gmail.com
}

\usepackage{bibentry}
\begin{document}

\maketitle

\begin{abstract}
Mixup is a data augmentation technique that enhances model generalization by interpolating between data points using a mixing ratio $\lambda$ in the image domain. Recently, the concept of mixup has been adapted to the graph domain through \emph{node-centric} interpolations. However, these approaches often fail to address the complexity of interconnected relationships, potentially damaging the graph's natural topology and undermining node interactions. Furthermore, current graph mixup methods employ an \emph{one-size-fits-all} strategy with a randomly sampled $\lambda$ for all mixup pairs, ignoring the diverse needs of different pairs. This paper proposes an Adaptive Graph Mixup (\textbf{\agmixup}) framework for semi-supervised node classification. \agmixup introduces a subgraph-centric approach, which treats each subgraph similarly to how images are handled in Euclidean domains, thus facilitating a more natural integration of mixup into graph-based learning. We also propose an adaptive mechanism to tune the mixing ratio $\lambda$ for diverse mixup pairs, guided by the contextual similarity and uncertainty of the involved subgraphs. Extensive experiments across seven datasets on semi-supervised node classification benchmarks demonstrate \agmixup's superiority over state-of-the-art graph mixup methods. Source codes are available at: \url{https://github.com/WeigangLu/AGMixup}.
\end{abstract}

\section{Introduction}
\label{sec:intro}
Graph Neural Networks (GNNs) have emerged as a dominant paradigm for learning on graph-structured data, driving significant advances in various graph-based tasks, notably node classification. Despite these successes, a fundamental challenge remains in how to effectively generalize from limited or sparsely labeled data, a common scenario in node classification. Mixup~\cite{mixup}, an effective data augmentation method originally devised for Euclidean data, enriches the training dataset by linearly interpolating between pairs of labeled samples (a mixup pair) and their corresponding labels with a mixing coefficient $\lambda$. While mixup has achieved notable success in Euclidean data domains characterized by regularity, its application to the complex, non-Euclidean nature of graph data encounters two significant questions.

\begin{figure}[!t]
\centering
\includegraphics[width=0.9\columnwidth]{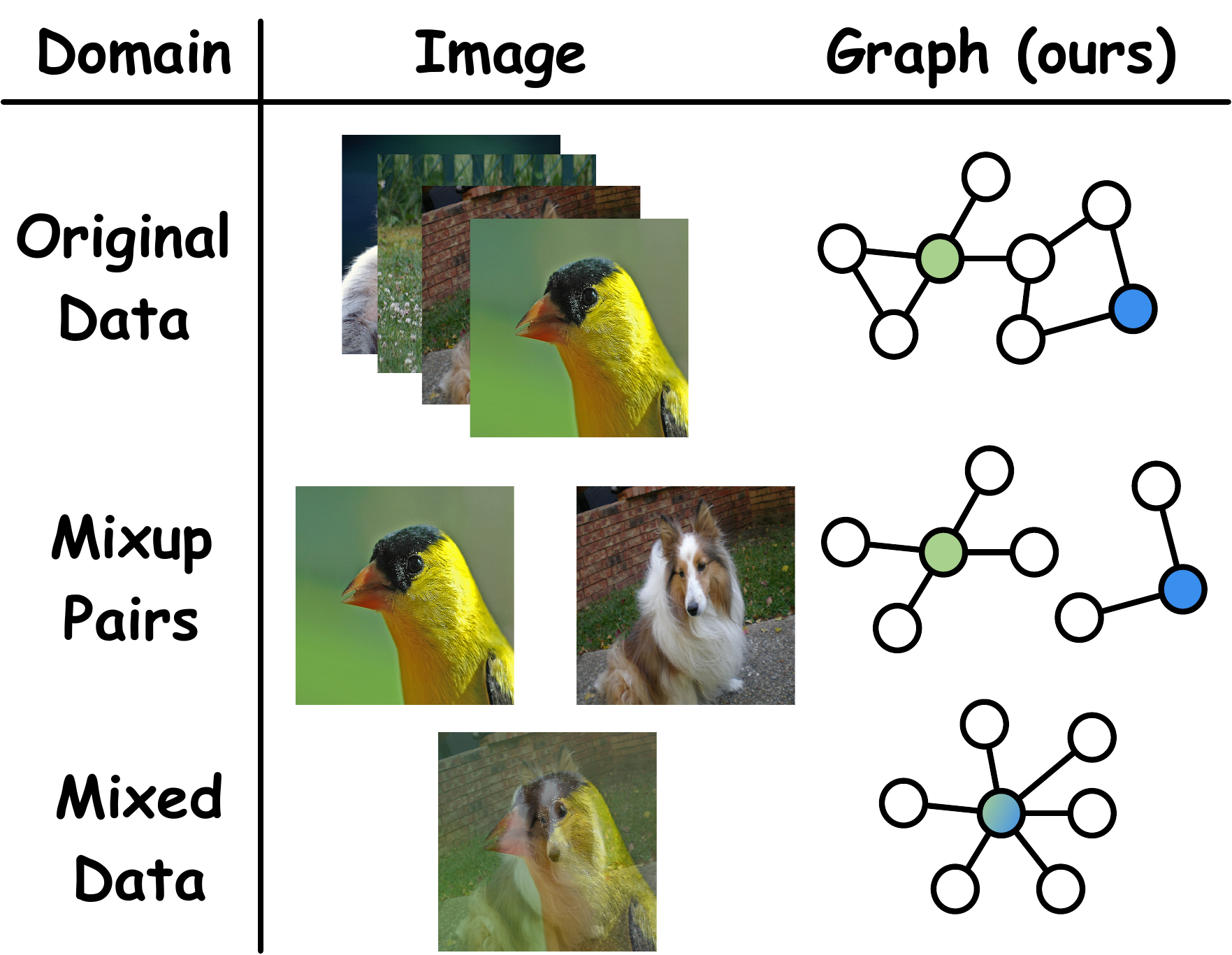}
\caption{Seamlessly integrating mixup from image domain into graph domain with our \agmixup.}
\label{fig:intro}
\end{figure}

\emph{Question 1: How can the mixup concept be seamlessly integrated into graphs?} Existing graph mixup methods, focusing on node-centric interpolations within a graph, extend the mixup method from Euclidean spaces to graphs by blending features, labels, and connections of node pairs~\cite{nodemixup}. However, these methods overlook the complexity of interconnected relationships, potentially altering the graph's natural topology and undermining the node interactions. In contrast to the isolation of images in mixup as shown in Fig.~\ref{fig:intro} (\textit{\textbf{left}}), where interpolations do not influence other data points, graph data's interconnected nature demands an approach to preserve its topological integrity. 

\emph{Question 2: How can the mixing ratio ($\lambda$) be adaptively controlled for different mixup pairs? } Prevailing graph mixup techniques often employ an one-size-fits-all strategy to control the mixing ratio, applying a randomly sampled $\lambda$ across all mixup operations. This approach can inadvertently generate synthetic nodes that starkly contrast with realistic distributions within the graph, especially when merging highly dissimilar node pairs with a mid-range $\lambda$. Moreover, this static strategy fails to encourage the model to explore less represented data areas, critical for more confident predictions. 

\begin{figure*}[!ht]
	 \centering
	 \includegraphics[width=0.8\linewidth]{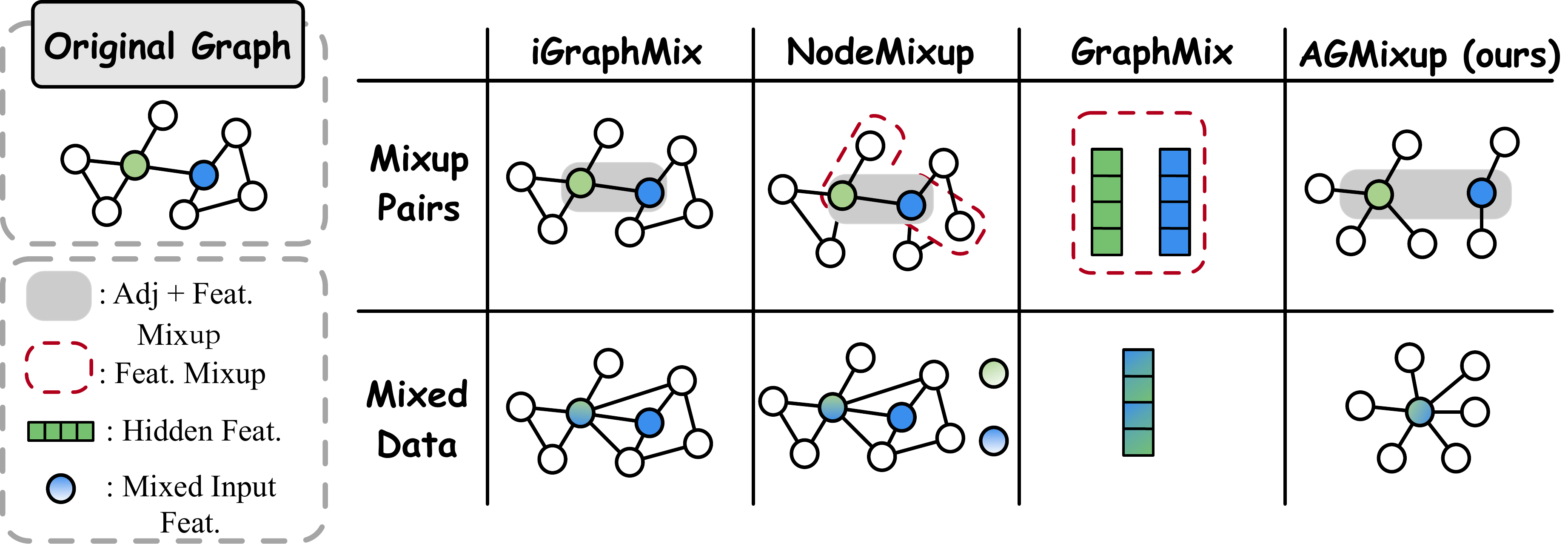}
     \caption{Difference between \agmixup and SOTA graph mixup methods.}
     \label{fig:discussion}
\end{figure*}

\textbf{Present Work.} In response to these challenges, we propose an \underline{A}daptive \underline{G}raph \underline{Mixup} (\textbf{\agmixup}) framework for semi-supervised node classification. \agmixup innovates a subgraph-centric mixup, extending the mixup concept beyond individual nodes. Given that a subgraph can encapsulate a node's local structure and semantic context, \agmixup treats a subgraph as analogous to an image, facilitating the transfer of the mixup idea from images to graphs, as shown in Fig.~\ref{fig:intro} (\textit{\textbf{right}}). This subgraph-centric approach not only mitigates the potential negative impacts of traditional in-graph mixup by treating each subgraph as an independent sample, but also offers scalability and flexibility in handling graphs of varying sizes. Furthermore, \agmixup introduces an adaptive mechanism for tuning the mixing ratio, guided by the contextual similarity and uncertainty associated with the subgraphs involved. This strategy ensures that $\lambda$ is optimally adjusted for diverse mixup pairs, leading to more realistic and exploratory synthetic graph data generation. Our main contributions are summarized as follows:
\begin{itemize}
\item We extend the mixup concept beyond the level of individual nodes to subgraphs. This methodology treats each subgraph as an analogous unit to an image in Euclidean domains, enabling a seamless integration of mixup into the graph domain.
\item We introduce an adaptive mechanism for dynamically tuning the mixing coefficient, based on the contextual similarity and uncertainty associated with the subgraphs involved. 
\item We have conducted extensive experiments demonstrating that \agmixup outperforms existing state-of-the-art (SOTA) graph mixup methods. Besides, we provide comprehensive empirical analysis to understand the behavior of \agmixup. 
\end{itemize}

\section{Related Work}
\label{sec:related_work}

\paragraph{Mixup on Images.}
Mixup~\cite{mixup}, introduced by Zhang \textit{et al.}, generates virtual training data by linearly interpolating the features and labels of image pairs using a ratio $\lambda$ sampled from a Beta distribution. This concept has been expanded into the hidden space by methods like ManifoldMixup~\cite{manifoldmixup} and PatchUp~\cite{patchup}, which apply interpolation to hidden representations to enhance sample diversity. Building upon this foundation, adaptive mixing policies such as PuzzleMix~\cite{puzzlemix}, SaliencyMix~\cite{saliencymix}, AutoMix~\cite{automix}, SuperMix~\cite{supermix}, and Decoupled Mixup~\cite{decoupledmixup} have been developed to refine the generation of mixed samples, tailoring the process to the specific characteristics of input data. Theoretical and empirical investigations~\cite{mixup_generaliztion,mixup_calibration} into mixup's effects on model generalization and calibration have further demonstrated its benefits, offering deep insights into mixup methods.

\paragraph{Mixup on Graphs.}
Extending mixup to the graph domain has opened new avenues for enhancing graph-level tasks, with several studies~\cite{ma2024fused,crisostomi2022metric,g_mixup,ifmixup_g,graphmad_g,mixup_transplant_g,mixup_metric_g} introducing mixup techniques to facilitate these tasks. For graph-level tasks, the integrity of the graph structure is not tied to specific nodes but to the overall graph. As a results, these methods might not scale efficiently when applied directly to node-level tasks since each node requires consideration of its unique neighborhood, which isn't necessarily the focus of whole-graph mixup methods. To address this challenge, \cite{mixup_for_node} proposes a method that mixes the input features of pairs of nodes and then blends their aggregated representations at each GNN layer. GraphMixup~\cite{graphmixup} incorporates a reinforcement mechanism to dynamically adjust the scale of mixup pairs, aiming to alleviate the class-imbalanced node classification problem. GraphMix~\cite{graphmix} applies mixup to the hidden representations of nodes and feeds the mixed nodes into fully connected layers. NodeMixup~\cite{nodemixup} introduces inter- and intra-class mixup techniques for both labeled and unlabeled nodes, addressing the issue of under-reaching. iGraphMix~\cite{igraphmix} focuses on mixing the features and connections of labeled nodes on the input graph.

\paragraph{Comparison to SOTA Graph Mixup.}
In Fig.~\ref{fig:discussion}, we illustrate the comparison between three SOTA graph mixup methods, i.e., iGraphMix, NodeMixup, GraphMix, and our \agmixup. Both iGraphMix and NodeMixup extend mixup operations to node connections within the graph, risking alterations to the graph's natural topology and potentially impacting node interactions negatively. Furthermore, these methods might introduce too many edges into graphs, easily inducing over-smoothing~\cite{dropedge,oono2020graph,skipnode}, particularly in large-scale graphs. The manifold mixup of GraphMix avoids this challenge by interpolating nodes in the latent space. However, this approach requires additional fully connected layers and may result in a loss of structural information. Besides, its flexibility is limited since hidden representations are not always accessible for linear GNNs (e.g., SGC~\cite{sgc}) either determinable for deep GNNs with layers larger than 2. Furthermore, all of these methods employ randomly sampled $\lambda$ values for all mixup pairs, disregarding the distinct needs of different mixup pairs. Our \agmixup addresses these limitations in the following ways: (1) employing subgraph-centric mixup, treating each subgraph as an independent sample; (2) introducing an adaptive mechanism to tune the mixing ratio for diverse mixup pairs.

\section{Methodology}
\label{sec:method}
In this section, we begin by revisiting Graph Neural Networks (GNNs) and the mixup technique. Subsequently, we introduce our proposed method, \agmixup, which comprises three key modules: subgraph-centric mixup, contextual similarity-aware $\lambda$ initialization, and uncertainty-aware $\lambda$ adjustment. Finally, we present a comprehensive case study to evaluate the effectiveness of \agmixup and analyze its computational complexity.

\subsection{Preliminaries}
\paragraph{Notations.}
We define an undirected graph with self-loops, $\mathcal{G}=\{\mathcal{V}, \mathcal{E}\}$, where $\mathcal{V}=\{\aVec{x}_{1}, \cdots, \aVec{x}_{N}\}$ represents the set of $N$ nodes, and $\mathcal{E}$ denotes the edge set with element $e_{ij}$ indicating a connection between node $i$ and node $j$. The input feature matrix $X \in \mathbb{R}^{N \times F}$, with its $i$-th row vector $\aVec{x}_{i}$, where $F$ represents the input dimensionality. Besides, the label matrix $Y \in \mathbb{R}^{N \times C}$, where $C$ is the number of classes, includes $\aVec{y}_{i}$ as the one-hot encoding of node $i$'s label. 

%Specifically, we divide the data set into labeled set $\labelset = \{\mathcal{V}_{L}, Y_{L}\}$ and unlabeled set $\unlabelset = \{\mathcal{V}_{U}, Y_{U}\}$.

\paragraph{GNN Revisit.}
Given a node $i$ with its feature vector $\aVec{x}_{i}$, a GNN computes the node's embedding $\aVec{h}_{i}$ through a series of layer-wise propagation rules. At each layer $l$, the embedding $\aVec{h}^{(l)}_{i}$ is updated as 
$
	\aVec{h}^{(l)}_{i+1} = \sigma \left(W^{(l)} \cdot \text{AGGREGATE} (\{\aVec{h}^{(l)}_{i} | j \in \mathcal{N}_{i}\})\right),
$
where $\sigma$ is a non-linear activation function, $W^{(l)}$ is the learnable parameter of $l$-th layer, and $\text{AGGREGATE}$ is a function that combines the embeddings of node $i$'s neighbors $\mathcal{N}_{i}$. The initial embedding $\aVec{h}^{(0)}_{i}$ is set to the node features, i.e., $\aVec{h}^{(0)}_{i} = \aVec{x}_{i}$. Upon completion of $L$ layers of propagation, the final embedding $\aVec{h}^{(L)}_{i}$ is then passed through a classification layer to obtain the predictions for each node 
$
	\aVec{p}_{i} = \text{softmax} (W^{(L)}\aVec{h}^{(L)}_{i}),
$
where parameters $W^{(L)}$ maps the final embedding into the prediction space and the softmax function ensures the output $\aVec{p}_{i}$ represents a probability distribution over the $C$ classes.
 
\paragraph{Mixup Revisit.}
The traditional mixup method is a data augmentation technique that creates virtual training samples by linearly interpolating between pairs of examples. Given two mixup candidates $\aVec{x}_{i}$ and $\aVec{x}_{j}$ from the labeled data $\data$, the mixup data $\mixpoint{x}{ij}$ can be generated as follows:
\begin{small}
	\begin{equation}
	\label{eq:mixup}
	\mixpoint{x}{ij} = \mathcal{M}(\aVec{x}_{i}, \aVec{x}_{j}, \lambda) = \lambda \aVec{x}_{i} + (1-\lambda) \aVec{x}_{j},
\end{equation}
\end{small}where $\mathcal{M}(.)$ is a mixup function. Here, $\lambda$ is a mixing coefficient drawn from a Beta distribution $\betadist(\alpha, \alpha)$ with $\alpha > 0$. The value of $\lambda$ controls the degree to which they are mixed. In the context of a general classification task, we regard the mixed data as additional training data with mixed labels. 

\subsection{Proposed Method: AGMixup}
\label{sec:agmixup}
\paragraph{Subgraph-Centric Mixup.}
Inspired by image mixup techniques, we treat subgraphs induced from nodes as independent samples since a subgraph encapsulates a node's local structure and semantic context. For a node $i$, its $r$-ego graph $\mathcal{G}^{(r)}_{i} =\{\mathcal{V}^{(r)}_{i}, \mathcal{E}^{(r)}_{i}\}$ includes all nodes and edges within $r$ hops. Our \agmixup innovatively proposes the mixing of such subgraphs, yielding a mixed graph $\mixpoint{\mathcal{G}}{ij}$ formulated as:
\begin{small}
	\begin{equation}
	\label{eq:gmixup}
	\mixpoint{\mathcal{G}}{ij} = \mathcal{M}_{g}(\mathcal{G}^{(r)}_{i}, \mathcal{G}^{(r)}_{j}, \lambda) = \{\mixpoint{\mathcal{V}}{ij}^{(r)}, \mixpoint{\mathcal{E}}{ij}^{(r)}\},
\end{equation}
\end{small}where $\mathcal{M}_{g} (.)$ is a mixup function for the graph. The mixed node set $\mixpoint{\mathcal{V}}{ij}^{(r)}$ contains all the nodes from both $\mathcal{G}^{(r)}_{i}$ and $\mathcal{G}^{(r)}_{j}$, excluding $\aVec{x}_{i}$ and $\aVec{x}_{j}$. Instead, a virtual node $\mixpoint{x}{ij} = \mathcal{M}(\aVec{x}_{i}, \aVec{x}_{j}, \lambda_{ij})$ is introduced, where $\lambda_{ij}$ is dynamically computed through our adaptive mechanism (detailed later). The mixed edge set $\mixpoint{\mathcal{E}}{ij}^{(r)}$ includes connections from the virtual node $\mixpoint{x}{ij}$ to the neighbors of both $\aVec{x}_{i}$ and $\aVec{x}_{j}$. It preserves the structural connectivity inherent to each original subgraph but also facilitates the integration of node features from the two distinct subgraphs. By doing so, \agmixup maintains the graph’s structural integrity and leverages the combined features to enhance learning efficacy, mirroring how traditional image mixup enhances model generalization by interpolating between distinct image samples.

\paragraph{Contextual Similarity-aware $\lambda$ Initialization.}
A fixed $\lambda$ drawn from $\betadist(\alpha, \alpha)$ applies the same degree of interpolation across all sample pairs, potentially limiting the model's ability to generalize from the mixed samples without an appropriate $\alpha$. A too small $\alpha$ predominantly yields $\lambda$ values near the extremes (close to one or zero), generating virtual samples overly similar to original samples. On the contrary, a larger $\alpha$ biases $\lambda$ towards 0.5. However, such uniform blending could result in unrealistic feature combinations when mixing highly dissimilar samples, diverging from the natural data distribution. To this end, we propose a similarity-aware $\lambda$ initialization mechanism:
\begin{small}
\begin{equation}
	\label{eq:lambda-sim}
	\lambda_{ij}^{(0)} = 0.5 * \exp(-\gamma \lVert \bar{\aVec{h}}^{(r)}_{i} - \bar{\aVec{h}}^{(r)}_{j} \rVert^{2}),
\end{equation}
\end{small}where $\bar{\aVec{h}}^{(r)}_{i}$ is the mean encoded embedding vector by a GNN model of all the nodes in $\mathcal{G}^{(r)}_{i}$ and $\gamma > 0$ adjusts the sensitivity of mixup to contextual similarity. Eq.~(\ref{eq:lambda-sim}) enables more exploratory mixing only for subgraphs that are similar in hidden space by initializing $\lambda_{ij}$ value close to 0.5. 

\paragraph{Uncertainty-aware $\lambda$ Adjustment.}
During the training with labeled nodes sparsely distributed in the graph, some regions of the feature space might be underrepresented, leading to higher uncertainty in those areas. An adaptive mixup strategy that accounts for uncertainty can ensure that mixed samples from these underrepresented areas are more frequently generated, contributing meaningfully to the learning process. An adaptive $\lambda$ adjustment, biased towards the node with higher uncertainty, can help the model focus on learning from the underrepresented data, improving its predictive performance across a more diverse set of nodes. As training progresses and the model becomes more confident in certain regions of the feature space (low uncertainty), adjusting $\lambda_{ij}$ to favor less mixing (closer to 0 or 1) for these samples encourages exploitation. This refinement phase allows the model to learn representations that generalize better to unseen data, enhancing the confidence of models. Based on this, we adjust $\lambda_{ij}$ based on the uncertainties of mixup candidates:
\begin{small}
\begin{equation}
	\label{eq:lambda-un}
	\lambda_{ij} = \lambda_{ij}^{(0)} + \Delta\lambda_{ij} = \lambda_{ij}^{(0)} + \beta (\frac{\bar{u}_{i} - \bar{u}_{j}}{U_{max}}),
\end{equation}
\end{small}where $\beta$ is a scaling factor that controls the influence of uncertainty difference on $\lambda_{ij}$. Given the mean  predicted probability vector $\bar{p}^{(r)}_{i}$ for all the nodes from $\mathcal{G}^{(r)}_{i}$, the uncertainty $\bar{u}_{i}$ is calculated as:
$
	\bar{u}_{i} = -\sum_{c}^{C} \bar{p}^{(r)}_{ic} \log \bar{p}^{(r)}_{ic},
$
where $\bar{p}^{(r)}_{ic}$ is the predicted probability over class c. Here, $U_{max} = \log C$ is a normalization term representing the maximum possible uncertainty difference, ensuring that the adjustment factor falls within a reasonable range. To ensure the calculated $\lambda_{ij}$ is within valid bounds, we apply a clipping operation on it: $\lambda_{ij} = \text{Clip}(\lambda_{ij}, 0, 1)$.

\paragraph{Training with AGMixup.}
In node classification tasks, the relationship between graph data $\mathcal{G}$ and node labels $\{\aVec{y}_{1}, \cdots, \aVec{y}_{N} \}$ is modeled through a graph neural network (e.g., GCN), denoted as $g_{\theta}: \mathcal{G} \mapsto \{\aVec{y}_{1}, \cdots, \aVec{y}_{N} \}$, with $\theta$ representing the network parameters. The optimization is guided by minimizing $\mathcal{L}_{G}$ as follows:
\begin{small}
\begin{equation}
	\label{eq:gmixup-ce}
	 \mathcal{L}_{G} = \frac{1}{|\mathcal{V}_{L}|}\ell(g_{\theta}(\mathcal{G}), \aVec{y}), 
\end{equation}
\end{small}where $\ell$ is the cross-entropy loss. To regularize the training process and encourage the model to learn from these mixed graphs effectively, \agmixup employs a mixup loss function:
\begin{small}
\begin{equation}
	\begin{split}
	\label{eq:gmce}
	\mathcal{L}_{M} 
			&= \frac{1}{|\mathcal{P}|}\sum_{(i,j) \in \mathcal{P}} \tilde{\ell}(\mixpoint{\mathcal{G}}{ij}, \aVec{y}_{i}, \aVec{y}_{j})\\
			& = \frac{1}{|\mathcal{P}|}\sum_{(i,j) \in \mathcal{P}} \lambda_{ij} \ell(g_{\theta}(\mixpoint{\mathcal{G}}{ij}), \aVec{y}_{i})  + (1 - \lambda_{ij}) \ell(g_{\theta}(\mixpoint{\mathcal{G}}{ij}), \aVec{y}_{j}),
	\end{split}
\end{equation}
\end{small}where $\mathcal{P}$ is the mixup pair index set. \emph{In practice, AGMixup generates a virtual graph $\tilde{\mathcal{G}}$ that encompasses a set of virtual disjoint subgraphs, each subgraph $\mixpoint{\mathcal{G}}{ij}$ corresponding to a mixup between pairs of labeled-node-induced graphs from the original graph.} Then, we combine Eq.~(\ref{eq:gmixup-ce}) and Eq.~(\ref{eq:gmce}) with a balance factor $\mu \in [0, 1]$ as follows:
\begin{small}
\begin{equation}
	\label{eq:overall-agmixup}
	\mathcal{L} = \mathcal{L}_{G} + \mu \mathcal{L}_{M}.
\end{equation}
\end{small}This mixup loss $\mathcal{L}_{M}$ serves as a regularization term to be added to the classification loss, guiding the model to not only fit the original data but also to generalize across the mixed data samples. By adaptively tuning the mixing parameter for each mixup pair, \agmixup enhances the model's interpolation capability and confidence. The algorithm of our \agmixup is provided in Appendix.

\subsection{Case Study: What is AGMixup doing?}
\label{sec:case_study}
In this section, we conduct a comprehensive empirical investigation into \agmixup on Cora, Pubmed, and Coauthor CS datasets. We benchmark \agmixup against various methodologies, including two variants of \agmixup itself (\textit{R}-GMixup, which randomly samples $\lambda$ from a Beta distribution, and \textit{F}-GMixup, employing a fixed $\lambda=0.5$ for all mixup pairs), two prevalent graph mixup methods (GraphMix and iGraphMix), and a baseline vanilla GCN model.

 \begin{figure}[!hbpt]
	 \centering
	 \subfigure{
	 			\includegraphics[width=0.47\columnwidth]{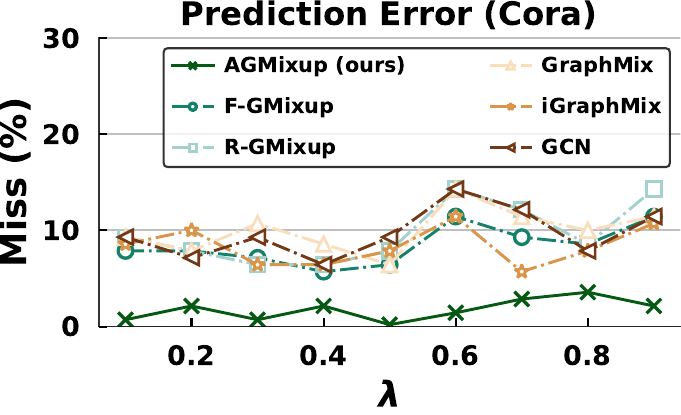}
		}
	\subfigure{
	 			\includegraphics[width=0.47\columnwidth]{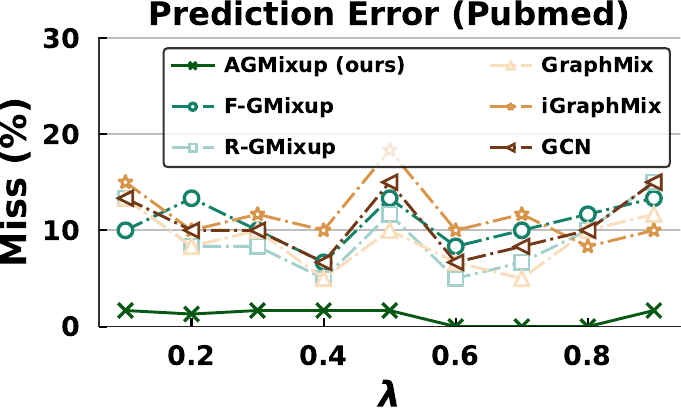}
		}
%	\hfil
%	\subfigure{
%	 			\includegraphics[width=0.3\linewidth]{figs/pred_erro-CS.pdf}
%		}
     \caption{Prediction errors in-between training data. Our \agmixup significantly reduces the rate of prediction misses, indicating superior interpolation capability.}
     \label{fig:pred_erro}
\end{figure}

\paragraph{Improving Interpolation Capability.}
Mixup~\cite{mixup} encourages models to behave linearly in-between training samples. In the context of the mixup method and its evaluation, a ``miss'' refers to an instance where the prediction for a mixed sample does not match either of the labels of the original samples. Fewer misses imply that the model is better at interpolating between classes, providing smooth transitions, and understanding the variations in data that might not be explicitly represented in the training set. In Fig.~\ref{fig:pred_erro}, we synthesize test subgraphs $\mixpoint{\mathcal{G}}{ij}$ with various $\lambda$. While all the baselines exhibit higher misses around the $\lambda=0.5$ point, our \agmixup shows a noticeable reduction in misses. For example, on the Cora dataset, \agmixup dramatically reduces the prediction misses to as low as 0.20\%, a stark contrast to the other methods, where misses even exceed 10\%. It demonstrates \agmixup's superior ability to enhance the model's ability to interpolate between training examples, thereby improving its interpolation capability. By tuning the mixing parameter adaptively, our \agmixup tailors the generation of mixed samples in a way that aligns more closely with the inherent data distribution, effectively reducing the incidence of prediction misses. 

 \begin{figure}[!ht]
	 \centering
	 \subfigure{
	 			\includegraphics[width=0.47\columnwidth]{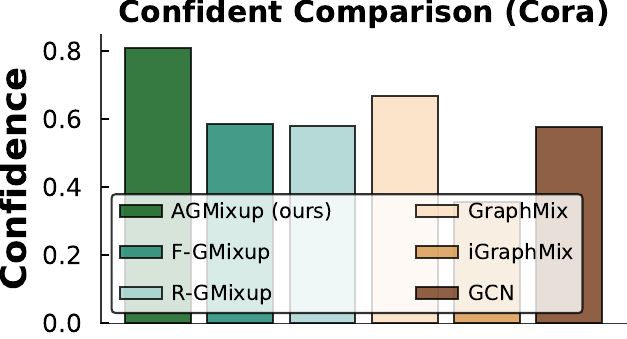}
		}
	\subfigure{
	 			\includegraphics[width=0.47\columnwidth]{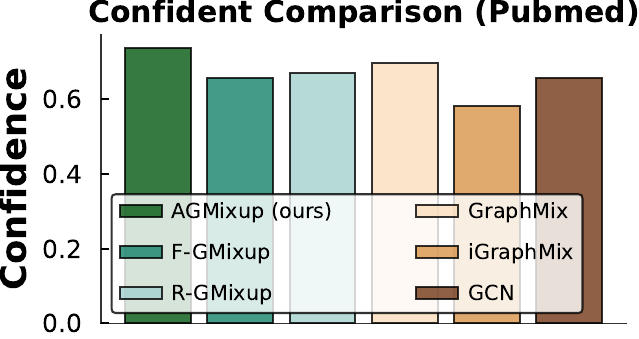}
		}
%	\hfil
%	\subfigure{
%	 			\includegraphics[width=0.3\linewidth]{figs/conf-CS.pdf}
%		}
     \caption{Confidence comparison. Our \agmixup makes the model more ``confident'' in the predictions by paying more attention to those underrepresented samples.}
     \label{fig:conf}
\end{figure}

\paragraph{Improving Model Confidence.}
Graphs often contain underrepresented regions when labeled nodes distribute sparsely in the graph~\cite{nodemixup}, which can lead to higher uncertainty in model predictions for those areas. Our \agmixup offers a dynamic balance between exploring new areas of the feature space (especially those that are underrepresented and uncertain) and exploiting the knowledge that the model has confidently acquired. As the model becomes more confident in certain regions (indicating low uncertainty), the adaptive mechanism adjusts to encourage less mixing for those samples, shifting towards exploitation of the learned representations. This approach encourages the model to pay more attention to underrepresented feature space regions. Empirical observations from Fig.~\ref{fig:conf} show that \agmixup significantly improves the confidence\footnote{The confidence is calculated by averaging the maximum prediction scores on nodes.} of GCN for test node. For instance, on the Cora dataset, \agmixup's confidence score surpasses 0.8, markedly higher than any baselines, suggesting that it successfully directs the model's focus towards previously underexplored areas, thereby enhancing overall prediction certainty. 

\begin{table*}[ht!]
    \centering
    
%    \resizebox{0.9\linewidth}{!}{
    \begin{tabular}{l c c c c c c  c}
    \hline
    \toprule
%    ~ & \multicolumn{3}{c}{\textbf{Medium Graphs}} & \multicolumn{3}{c}{\textbf{Large Graphs}} & \multirow{4}{*}{\textbf{Avg. Gains}} \\
    
    ~ & Cora & Citeseer & Pubmed & CS & Physics & Arxiv & \textbf{Avg. Gains} \\
%    \midrule
%	\cmidrule(r){2-7}  
	
%	\textbf{\# Nodes} & \emph{2.7k} & \emph{3.3k} & \emph{19.7k} & \emph{18.3k} & \emph{34.4k} & \emph{169.3k} & ~\\
%	\textbf{\# Edges} & \emph{5.2k} & \emph{4.6k} & \emph{44.3k} & \emph{81.8k} & \emph{247.9k}2 & \emph{1166.2k} & \textbf{Avg. Gains}\\

%	\cmidrule(r){1-1} \cmidrule(r){2-7} \cmidrule(r){8-8} 
	\midrule
	GCN  
    & 80.51$_{\pm1.39}$ 
    & 69.53$_{\pm0.96}$ 
    & 78.23$_{\pm1.64}$ 
    & 91.22$_{\pm0.83}$ 
    & 93.51$_{\pm0.34}$ 
    & 65.39$_{\pm0.13}$ 
    & \zero \\
    \hdashline[5pt/4pt]
    ~ + GraphMix 
    & 81.18$_{\pm1.03}$ 
    & 69.67$_{\pm0.96}$ 
    & 78.07$_{\pm1.99}$ 
    & 91.27$_{\pm0.97}$ 
    & 93.54$_{\pm0.73}$ 
    & 65.27$_{\pm0.29}$ 
    & \imp{0.12} \\
    
    ~ + NodeMixup 
    & 81.21$_{\pm1.02}$ 
    & 69.45$_{\pm0.74}$ 
    & 78.98$_{\pm1.32}$ 
    & 92.03$_{\pm0.36}$ 
    & 93.86$_{\pm0.67}$ 
    & 66.82$_{\pm0.24}$ 
    & \imp{0.86}\\
    
    ~ + iGraphMix 
    & 81.16$_{\pm0.95}$ 
    & 69.95$_{\pm0.84}$ 
    & 77.07$_{\pm1.61}$ 
    & 90.93$_{\pm0.68}$ 
    & 93.61$_{\pm0.36}$ 
    & 65.97$_{\pm0.57}$ 
    & \imp{0.10}\\
    
    ~ + \textbf{\agmixup (\textit{ours})}
    & \textbf{83.61}$_{\pm0.39}$ & \textbf{70.63}$_{\pm1.54}$ & \textbf{81.90}$_{\pm0.40}$ & \textbf{92.35}$_{\pm0.92}$ & \textbf{94.45}$_{\pm0.42}$ & \textbf{67.96}$_{\pm0.21}$ & \imp{\textbf{2.71}}\\ 
    
    \midrule
%	\cmidrule(r){1-1} \cmidrule(r){2-4}  \cmidrule(r){5-7} \cmidrule(r){8-8}
	    
    GAT 
    & 80.02$_{\pm1.15}$ 
    & 68.38$_{\pm1.35}$ 
    & 77.28$_{\pm1.80}$ 
    & 90.06$_{\pm0.80}$ 
    & 92.59$_{\pm0.64}$ 
    & 66.57$_{\pm0.27}$ & 
    \zero\\
    \hdashline[5pt/4pt]
    ~ + GraphMix 
    & 80.07$_{\pm1.11}$ 
    & 69.53$_{\pm0.76}$ 
    & 76.96$_{\pm1.85}$ 
    & 88.43$_{\pm1.49}$ 
    & 91.73$_{\pm1.47}$ 
    & 65.84$_{\pm0.18}$ 
    & \dec{0.41}\\
    
    ~ + NodeMixup 
    & 81.46$_{\pm1.13}$ 
    & 69.21$_{\pm0.86}$ 
    & 78.03$_{\pm1.03}$ 
    & 90.86$_{\pm1.23}$ 
    & 92.84$_{\pm1.42}$ 
    & 66.72$_{\pm0.26}$ 
    & \imp{0.89}\\
    
    ~ + iGraphMix 
    & 80.07$_{\pm1.20}$ 
    & 69.11$_{\pm0.79}$ 
    &77.14$_{\pm1.35}$ 
    & 90.01$_{\pm0.99}$ 
    & 92.64$_{\pm0.69}$ 
    & 64.79$_{\pm0.51}$ 
    & \dec{0.28}\\
    
    ~ + \textbf{\agmixup (\textit{ours})}
    & \textbf{83.14}$_{\pm0.54}$ & \textbf{69.72}$_{\pm1.51}$ & \textbf{81.60}$_{\pm0.51}$ & \textbf{91.92}$_{\pm1.41}$ & \textbf{94.31}$_{\pm0.63}$  & \textbf{67.48}$_{\pm0.72}$ & \imp{\textbf{2.78}}\\ 

    \midrule
%	\cmidrule(r){1-1} \cmidrule(r){2-4}  \cmidrule(r){5-7} \cmidrule(r){8-8}
	    
    JKNet 
    & 79.48$_{\pm1.47}$ 
    & 67.45$_{\pm1.05}$ 
    & 77.52$_{\pm2.29}$ 
    & 89.93$_{\pm1.11}$ 
    & 92.59$_{\pm1.36}$ 
    & 67.39$_{\pm0.18}$ 
    & \zero\\
     \hdashline[5pt/4pt]
    ~ + GraphMix 
    & 80.14$_{\pm1.29}$ 
    & 67.75$_{\pm1.01}$ 
    & 78.11$_{\pm2.12}$ 
    & 90.97$_{\pm0.85}$ 
    & 92.68$_{\pm1.09}$ 
    & 66.59$_{\pm0.13}$ 
    & \imp{0.35}\\
    
    ~ + NodeMixup 
    & 80.94$_{\pm1.03}$ 
    & \textbf{68.08}$_{\pm1.41}$ 
    & 78.96$_{\pm1.61}$ 
    & 91.71$_{\pm0.98}$ 
    & 92.44$_{\pm1.24}$ &
     67.98$_{\pm0.42}$ 
    & \imp{1.22}\\
    
    ~ + iGraphMix 
    & 79.94$_{\pm1.03}$ 
    & 67.58$_{\pm0.99}$ 
    & 76.03$_{\pm2.10}$ 
    & 91.19$_{\pm0.58}$ 
    & 92.83$_{\pm0.86}$ 
    & 65.15$_{\pm0.81}$ 
    & \dec{0.46}\\
    
    ~ + \textbf{\agmixup (\textit{ours})} 
    & \textbf{82.06}$_{\pm0.83}$ & 68.00$_{\pm1.53}$ & \textbf{81.34}$_{\pm0.44}$  & \textbf{92.63}$_{\pm0.50}$ & \textbf{93.27}$_{\pm0.62}$ & \textbf{69.76}$_{\pm0.21}$ & \imp{\textbf{2.70}}\\

    \midrule
%	\cmidrule(r){1-1} \cmidrule(r){2-4}  \cmidrule(r){5-7} \cmidrule(r){8-8}
	    
    GraphSAGE 
    & 79.28$_{\pm1.20}$ 
    & 68.36$_{\pm0.89}$ 
    & 75.72$_{\pm1.84}$ 
    & 91.95$_{\pm0.36}$ 
    & 93.18$_{\pm0.47}$ 
    & 66.00$_{\pm0.17}$ 
    & \zero\\
    \hdashline[5pt/4pt]
    ~ + GraphMix 
    & 79.65$_{\pm1.44}$ 
    & 68.44$_{\pm1.21}$ 
    & 75.88$_{\pm1.89}$ 
    & 91.71$_{\pm0.52}$ 
    & 92.93$_{\pm0.59}$ 
    & 66.73$_{\pm0.14}$ 
    & \imp{0.22}\\

    ~ + NodeMixup 
    & 80.42$_{\pm1.23}$ 
    & 69.06$_{\pm0.58}$ 
    & 77.92$_{\pm1.72}$ 
    & 91.94$_{\pm0.38}$ 
    & 93.78$_{\pm0.68}$ 
    & 67.13$_{\pm0.21}$ 
    & \imp{1.28}\\
    
    ~ + iGraphMix 
    & 79.72$_{\pm1.14}$ 
    & 68.30$_{\pm0.82}$ 
    & 75.70$_{\pm1.67}$ 
    & 91.70$_{\pm0.61}$ 
    & 92.80$_{\pm0.77}$ 
    & 64.66$_{\pm0.47}$ 
    & \dec{0.37}\\
    ~ + \textbf{\agmixup (\textit{ours})} 
    & \textbf{82.67}$_{\pm1.16}$ & \textbf{70.04}$_{\pm1.08}$ & \textbf{79.84}$_{\pm0.74}$  & \textbf{92.31}$_{\pm0.50}$ & \textbf{94.28}$_{\pm0.64}$  & \textbf{68.82}$_{\pm0.11}$ & \imp{\textbf{3.00}}\\ 
    
    \bottomrule
%	\addlinespace
%    \multicolumn{8}{l}{\footnotesize{The best accuracy is marked in \textbf{bold} according to each backbone model. Results on ogbn-products are in the Appendix.}}
    \end{tabular}
%    }
    \caption{Semi-supervised Node Classification Results (\%).\label{tab:semi}}

\end{table*}

\subsection{Complexity Analysis}
\agmixup 's complexity is driven by three operations: subgraph extraction, mixup operation, and adaptive $\lambda$ calculation. Subgraph extraction can be done offline, minimizing training overhead. The mixup operation, which involves interpolating features and connections, is computationally light due to parallelizable matrix addition. We assume that each subgraph contains $n$ nodes and has a hidden dimensionality of $d$. Therefore, adaptive $\lambda$ calculation, with a complexity of $O(n + nd)$ for similarity and $O(n + nC)$ for uncertainty, remains manageable. To ensure scalability, we introduce a shrink ratio $\epsilon \in (0, 1]$ to control the number of subgraphs processed per iteration, keeping the overall complexity at $O(\epsilon|\labelset|(n+C+2))$, where $\labelset$ is the labeled node set.

\begin{figure}[!htp]
\centering
\includegraphics[width=\columnwidth]{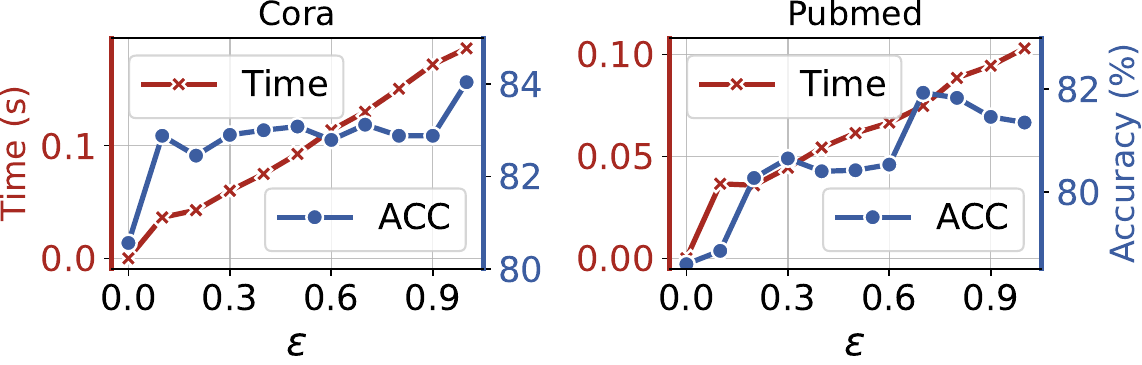}
\caption{Efficacy analysis using GCN as backbone model.} 
\label{fig:efficacy}
\end{figure}
In Fig.~\ref{fig:efficacy}, we analyze the relationship between the time cost per epoch of mixup training and accuracy of \agmixup w.r.t different values of $\epsilon$, ranging from 0 to 1 with an interval of 0.1.  The results show a clear trend: as $\epsilon$ increases, computational time also rises, reflecting the additional overhead associated with using a larger proportion of data for mixup. This increase in computational cost is expected as more subgraphs are processed during training. On the other hand, model performance generally improves with higher $\epsilon$ values, peaking at certain points before plateauing or slightly declining. This suggests that while increasing $\epsilon$ enhances model generalization, there is an optimal range where performance gains are maximized without incurring excessive computational costs.
 
\section{Experiment}
\label{sec:exp}
We compare our \agmixup with three state-of-the-art graph mixup methods targeting at the semi-supervised node classification problem: (1) GraphMix~\cite{graphmix} that trains GNNs by interpolating nodes’ hidden representations and corresponding labels; (2) NodeMixup~\cite{nodemixup} that chooses mixup pairs from labeled and unlabeled node sets; (3) iGraphMix~\cite{igraphmix} that mixes the input features, labels, and connections of two labeled nodes. We choose four representative GNNs as our backbone models, i.e., GCN~\cite{gcn}, GAT~\cite{gat}, JKNet~\cite{jknet}, and GraphSAGE~\cite{sage}. As for the datasets, we choose seven graphs, i.e., Cora, Citeseer, Pubmed~\cite{cora}, Coauthor CS, and Coauthor Physics~\cite{coauthor}, and two large-scale graphs, i.e., ogbn-arxiv and ogbn-products~\cite{ogb}. 

\subsection{Comparison}
\label{sec:result}

\paragraph{Classification Accuracy.}
The semi-supervised node classification accuracy results presented in Table~\ref{tab:semi} are obtained from ten different runs, ensuring reliable and consistent measurements. The results consistently demonstrate that \agmixup surpasses both the baseline models and other mixup methods in terms of classification accuracy across all datasets and models. Notably, \agmixup exhibits significant performance improvements, achieving the highest enhancements in accuracy across almost all tested GNNs. For instance, when applied to the GCN model on Cora, \agmixup achieves a remarkable peak accuracy of 83.61\%, outperforming SOTA graph mixup methods by a significant margin. These notable improvements reveal the effectiveness of \agmixup in enhancing the model's ability to generalize from limited labeled data, as well as its versatility in being applied to various GNNs and datasets. The best accuracy is marked in \textbf{bold} according to each backbone model. 

\begin{figure}[!ht]
	 \centering
	 \subfigure{
	 			\includegraphics[width=0.47\columnwidth]{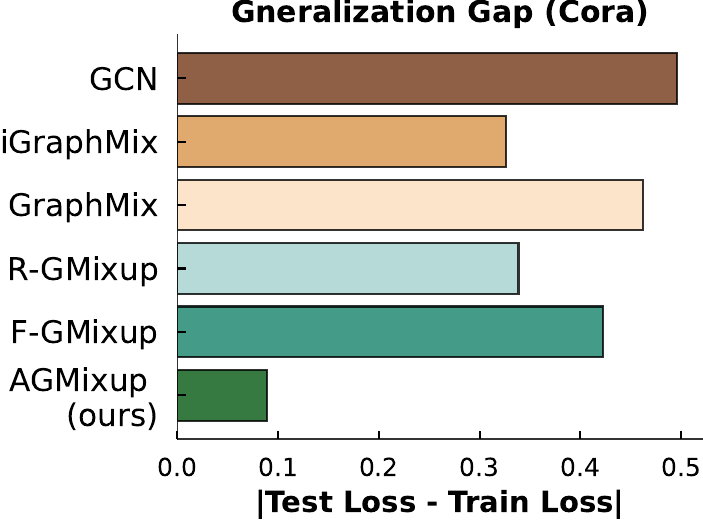}
		}
	\subfigure{
	 			\includegraphics[width=0.47\columnwidth]{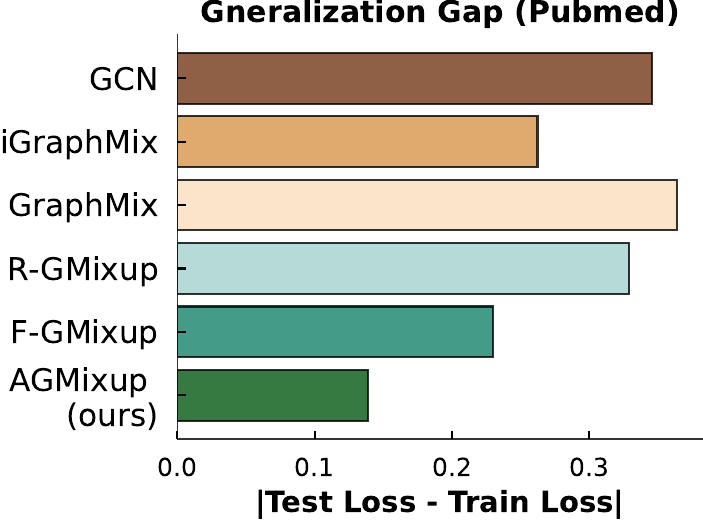}
		}
%	\hfil
%	\subfigure{
%	 			\includegraphics[width=0.3\linewidth]{figs/gap-CS.pdf}
%		}
     \caption{Generalization gap comparison. \agmixup shows better generalization ability over vanilla GCN and other methods, keeping the model from being over-fitting.}
     \label{fig:gap}
\end{figure}

\paragraph{Generalization Gap.}
We explore the generalizability of our \agmixup by calculating the generalization gap by comparing the generalization gaps across different methods in Fig.~\ref{fig:gap}. The generalization gap, defined as the difference between the test loss and training loss, serves as an indicator of a model's ability to generalize from training data to unseen test data. Lower values indicate better generalization, as the model is less likely to overfit the training data. The consistent outperformance of \agmixup across all datasets highlights its robustness and effectiveness in ensuring model generalization. The generalization advantage of \agmixup is due to its capacity to blend the strengths of mixup regularization with the specific challenges posed by graph data, creating a robust method that effectively bridges the gap between training and test performance.

\paragraph{Additional Results.}
Due to the page limit, we provide some results in Appendix: (1) a paired T-Test between \agmixup and other methods; (2) results on more advanced GNNs, i.e., GCNII~\cite{gcnii}, APPNP~\cite{appnp}, GPRGNN~\cite{gprgnn}, and GRAND~\cite{grand}; (3) results on fewer labeled nodes; (4) results on ogbn-products; (5) performance comparison against other graph augmentation methods, i.e., DropEdge~\cite{dropedge} and PairNorm~\cite{pairnorm}.

\subsection{Ablation Study}
\label{sec:ablation}
In this section, we ablate our \agmixup employing on GCN to investigate two important designs, i.e., subgraph-centric mixup and adaptive $\lambda$ generation.

\begin{figure}[!htp]
\setlength{\abovecaptionskip}{6pt}
\centering
\includegraphics[width=\columnwidth]{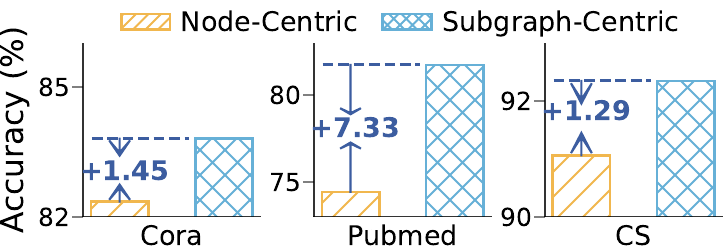}
\caption{Ablation study of subgraph-centric mixup.} 
\label{fig:abl_subgraph}
\end{figure}

\paragraph{Subgraph-Centric Mixup.}
In our comparative analysis illustrated in Fig.~\ref{fig:abl_subgraph}, we evaluate the performance of subgraph-centric against node-centric mixup across various graph datasets. The node-centric mixup, a variant of our \agmixup with $r=0$, focuses solely on the central nodes. The results clearly demonstrate the advantages of the subgraph-centric approach, especially on the Pubmed dataset where it significantly outperforms the node-centric version. The subgraph-centric mixup's superiority can be attributed to its capacity to retain more complex structural information and contextual dependencies within graphs. Unlike the node-centric approach, which concentrates on individual node features and may result in considerable information loss, subgraph-centric mixup preserves the integrity of local graph structures. This method not only captures the broader graph topology but also enhances model robustness by incorporating a richer contextual understanding into the training process. 

\begin{figure}[!htp]
\centering
\includegraphics[width=\columnwidth]{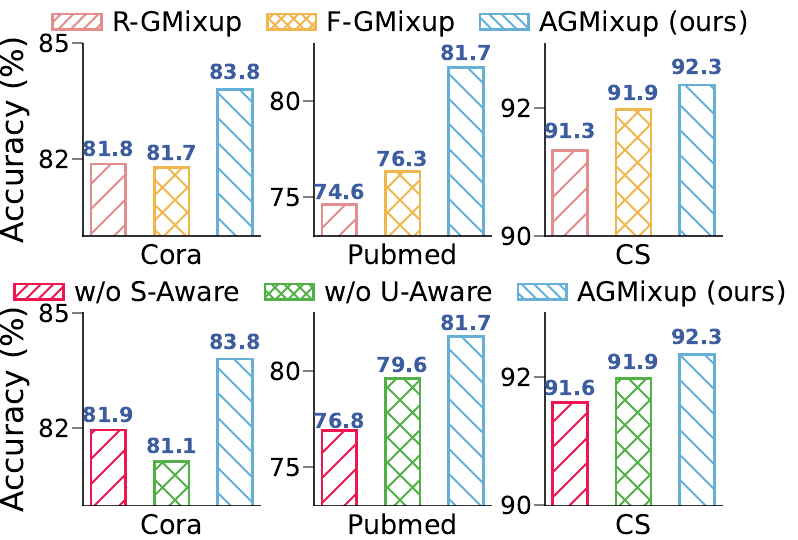}
\caption{Ablation study of adaptive $\lambda$ generation.} 
\label{fig:abl_lam}
\end{figure}

\paragraph{Adaptive $\lambda$ Generation.} 
Our analysis explores the effectiveness of adaptive $\lambda$ in \agmixup by contrasting it with \textit{R}-GMixup, which randomly samples $\lambda$ from a Beta distribution, and \textit{F}-GMixup, which employs a fixed $\lambda =0.5$. As illustrated in Fig.~\ref{fig:abl_lam} (\textbf{\textit{upper}}), \agmixup consistently outperforms these variants across all datasets. This superior performance underscores the significance of tailoring $\lambda$ to the distinct characteristics of each mixup pair, thus optimizing the integration of subgraphs for enhanced GNNs' performance. The second part of the analysis in Fig.~\ref{fig:abl_lam} (\textbf{\textit{bottom}}), evaluate the contributions of the Contextual Similarity-aware $\lambda$ Initialization (\textbf{S-Aware}) and the Uncertainty-aware $\lambda$ Adjustment (\textbf{U-Aware}) modules.  The removal of either module leads to a noticeable decrease in performance, with the impact being more severe in the absence of the S-Aware module. This is particularly evident in the Pubmed dataset, highlighting the S-Aware module's critical role in initializing $\lambda$ in response to the contextual similarity between subgraphs. This analysis confirms the critical importance of both modules in our \agmixup.

\subsection{Hyperparameter Study}
\label{sec:hyper}
In this section, we explore the impact of key hyperparameters on the performance of \agmixup, i.e., subgraph size (controlled by $r$) and sensitivity to similarity and uncertainty (controlled by $\gamma$ and $\beta$, respectively). We conduct experiments on Cora and Pubmed using GCN as the backbone model. 
 
\begin{figure}[!htp]
\centering
\includegraphics[width=\columnwidth]{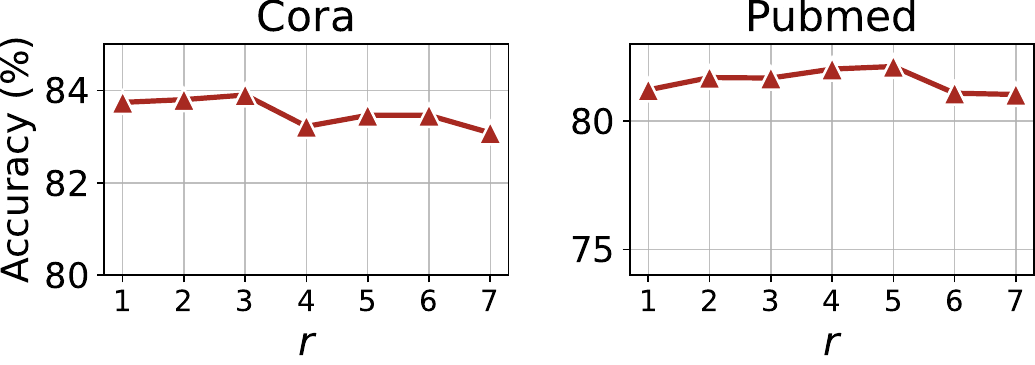}
\caption{Hyperparameter study on $r$.} 
\label{fig:hyp_subgraph_size}
\end{figure}

\paragraph{Subgraph Size.}
We investigate the effects of the subgraph size hyperparameter $r$ on the performance of \agmixup, ranging from 1 to 7, as illustrated in Fig.~\ref{fig:hyp_subgraph_size}. For the Cora dataset, \agmixup exhibits optimal performance at $r=3$, achieving the highest accuracy of 83.90\%, with performance declining as $r$ increased. For the Pubmed dataset, \agmixup shows improved performance as $r$ is increased up to a moderate level ($r=5$), peaking at an accuracy of 82.13\%. These trends suggest that while a larger $r$ can beneficially expand the contextual scope of the node features, excessively broad neighborhoods may introduce noise and irrelevant information. \emph{Based on these findings, to achieve an optimal balance between performance enhancement and computational efficiency, we recommend tuning $r$ within the range of 2 to 5.} 

\begin{figure}[!htp]
\centering
\includegraphics[width=\columnwidth]{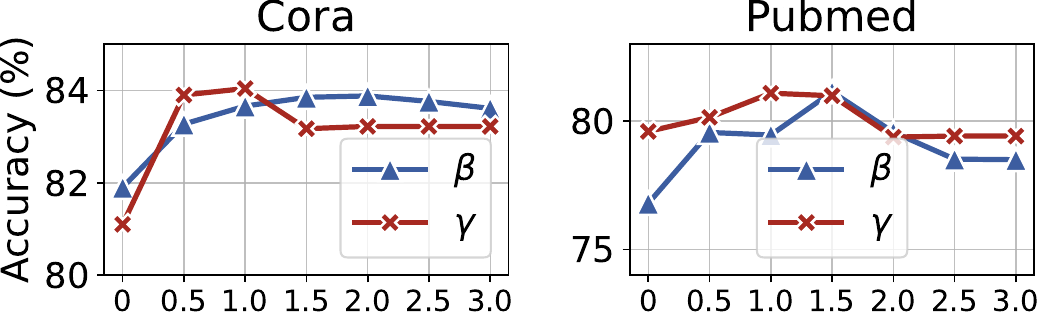}
\caption{Hyperparameter study on $\gamma$ and $\beta$.}
\label{fig:hyp_gamma_beta}
\end{figure}

\paragraph{Sensitivity to Similarity and Uncertainty.}
To evaluate the impact of $\gamma$ and $\beta$, which adjust the sensitivity to contextual similarity and the influence of uncertainty differences in mixup pairs respectively, we conduct a comprehensive analysis in Fig.~\ref{fig:hyp_gamma_beta}. The range of both hyperparameters is from 0 to 3 with an interval of 0.5. For both datasets, we observe that increasing $\beta$ initially improves performance.

 However, this trend reverses beyond an optimal point, suggesting that excessive emphasis on uncertainty might cause the model to focus too much on uncertain data, integrating too much randomness into the training process. Similarly, $\gamma$ also exhibits an initial increase in performance, reaching a peak before it begins to decline. This pattern indicates that while a moderate increase in $\gamma$ can help the model avoid generating unrealistic mixed data for dissimilar subgraphs, an overly high $\gamma$ potentially makes the mixup operation sensitive to dissimilarity. It limits the model's ability to generalize effectively to new, unseen data. \emph{Based on these empirical observations, we recommend setting $\gamma$ and $\beta$ at the range of 0.5 to 2.} This range appears to strike an effective balance between enhancing model performance.

\section{Conclusion and Future Work}
\label{sec:limitations}
Our \agmixup introduces significant advancements in applying adaptive mixup technique to the graph domain. Future work could focus on establishing a theoretical framework to understand how mixup methods affect GNNs and developing a generalized mixup method for different graph-based learning tasks, e.g., link prediction~\cite{linkpred} and graph classification~\cite{yang-graph-classification}.

\section{Acknowledgments}
This work was supported in part by the National Natural Science Foundation of China under Grants 62133012, 61936006, 62425605, and 62303366, the Fundamental Research Funds for the Central Universities under Grants QTZX24072, and the Key Research and Development Program of Shaanxi under Grant 2024CY2-GJHX-15.

\bibliography{aaai25}

\section*{Appendix}
In the Appendix section, we provide implementation details (\S\ref{appsec:impl}), the pytorch-like algorithm of \agmixup (\S\ref{appsec:alg}), and more results of experiment and empirical analysis (\S\ref{appsec:more_exp}).

\section{Algorithm}
\label{appsec:alg}
We provide the algorithm of our \agmixup in Algorithm~\ref{alg:agmixup}.

\input{algorithm/alg.tex}

\section{Implementation Details}
\label{appsec:impl}
\subsection{Datasets}
\label{appsec:dataset}
\begin{table*}[!htbp]
\begin{center}
%\vspace{-1.5em}
\caption{Datasets Statics.}
%\vspace{0.5em}
\label{tab:data_stat}
\resizebox{0.9\linewidth}{!}{
\begin{tabular}{c ccccc}
\toprule
~ & \textbf{Dataset} & \textbf{\# Nodes} & \textbf{\# Edges} & \textbf{\# Features} & \textbf{\# Classes}  \\ \midrule
\multirow{3}{*}{Medium Graphs}& Cora & 2,708 & 5,278 & 1,433 & 7  \\
~& Citeseer & 3,327 & 4,614 & 3,703 & 6  \\
~& Pubmed & 19,717 & 44,324 & 500 & 3  \\

\midrule

\multirow{4}{*}{Large Graphs} & Coauthor CS & 18,333 & 81,894 & 6,805 & 15 \\
~& Coauthor Physics & 34,493 & 247,962 & 8,415 & 5 \\
~& ogbn-arxiv & 169,343 & 1,166,243 & 128 & 40 \\ 
~& ogbn-products & 2,449,029 & 61,859,140 & 100 & 47 \\
\bottomrule
\end{tabular}} 
%\vspace{-1.5em}
\end{center}
\end{table*}

\paragraph{Dataset Description.}
We use 7 real-world graphs and provide statistics of these datasets in Table~\ref{tab:data_stat}. \underline{Cora, Citeseer, and Pubmed}\footnote{\url{https://linqs.soe.ucsc.edu/data}} are citation networks in which each publication is defined by a 0/1-valued word vector indicating the presence or absence of the corresponding dictionary word. \underline{Coauthor CS and Coauthor Physics}\footnote{\url{https://www.kdd.in.tum.de/gnn-benchmark}} are co-authorship graphs extracted from the Microsoft Academic Graph. Authors are represented as nodes, which are connected by an edge if two authors are both authors of the same paper. \underline{Ogbn-arxiv}\footnote{\url{https://github.com/snap-stanford/ogb}} is a paper citation network extracted from the Microsoft Academic Graph. The feature vector is generated for each paper (node) in the dataset by averaging the embeddings of words in the title and abstract. \underline{Ogbn-products}\footnotemark[4] is an e-commerce network characterized by product descriptions represented as nodes, and edges connecting nodes indicate that two products were purchased together. The node features in this network are constructed using a bag-of-words representation, and Principal Component Analysis (PCA) is applied to reduce the dimensionality of these node features.

\paragraph{Data Splits.}
 We randomly choose 20 nodes per class for training, 500 nodes for validation, and the rest for test on Cora, Citeseer, and Pubmed, 20 nodes per class for training, 30 nodes per class for validation, and the rest for test on Coauthor CS and Coauthor Physics. For ogbn-arxiv and ogbn-products, we abide by the dataset splits in the GitHub repository\footnotemark[4].

\subsection{Training Settings}
\label{appsec:training_setting}
\paragraph{Hardware and Software.}
\agmixup~is implemented based on the Torch Geometric library~\cite{torch_geo} and PyTorch 3.7.1 with Intel(R) Core(TM) i9-10980XE CPU @ 3.00GHz and 2 NVIDIA TITAN RTX GPUs with 24GB memory.

\paragraph{Hyperparameters Seaching.}
We set the maximum training epochs at 500 and implement early stopping after 50 epochs for all trials. For each backbone model, we set the hidden dimensionality at 128, the weight decay rate at 0.0005, and the learning rate at 0.01. We set the attention heads of GAT at 4 for ogbn-arxiv and 8 for the other datasets. The search space of the hyper-parameters is as follows:
\begin{itemize}
	\item Subgraph Size $r = 2$
	\item Sensitivity Parameters $\gamma, \beta = \{0.5, 1, 1.5, 2, 2.5, 3\}$ 
\end{itemize}
Regarding the balance factor between the classification loss and mixup loss $\mu$, we initially fix it at 1 while searching for other hyperparameters. Then, we tune $\mu$ using the values from $\{0.1, 0.3, 0.5, 0.7, 0.9\}$, while keeping the other hyperparameters frozen.

\section{More Experiment Results}
\label{appsec:more_exp}

%%%%%%%%%%% Significance Test %%%%%%%%%%%
\subsection{Significance Test}
\begin{figure*}[!ht]
	\includegraphics[width=0.97\linewidth]{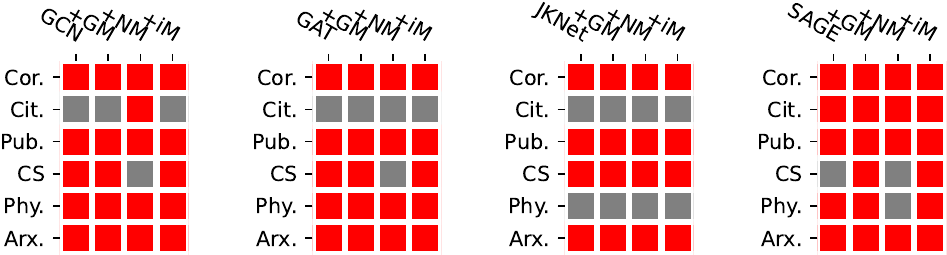}
	\caption{Paired T-Test between \agmixup and other mixup methods and backbone GNNs. In each subfigure, the first column represents the comparison between \agmixup and the backbone GNN, while the subsequent columns represent the comparisons between \agmixup and the corresponding mixup method ("GM": GraphMix, "NM": NodeMixup, "iM": iGraphMix). \textcolor{red}{Red} area represents $p$-value $<$ 0.05 for corresponding comparsion.}
	\label{appfig:ttest}
\end{figure*}

In Fig.~\ref{appfig:ttest}, we conduct a paired T-Test to statistically compare the performance of \agmixup against other mixup methods and the underlying backbone GNNs. This analysis is carried out over 10 independent runs for each model on each dataset. We denote statistically significant improvements, where the $p$-value $< 0.05$, in \textcolor{red}{red}. The results clearly demonstrate that \agmixup significantly outperforms both the conventional mixup methods and the baseline GNN models across most of the datasets tested. This robust performance highlights the effectiveness of \agmixup in enhancing model's performance.

%%%%%%%%%%% Results on Advanced GNNs %%%%%%%%%%%
\subsection{Results on Advanced GNNs}
\label{appsec:advanced_gnns}

%\begin{wraptable}[14]{r}{0.50\textwidth}
%\vspace{-0.4cm}
 \begin{table}[!t]
\caption{Results of \agmixup on advanced GNNs. }
\label{apptab:advanced_gnns}
\centering
\renewcommand\arraystretch{1.3} % 行间距
\setlength\tabcolsep{3pt} % 列间距
\resizebox{0.9\linewidth}{!}{%
\begin{tabular}{@{}lcccc@{}} %@{}:指定空格来消除列之间的间距，从而实现紧凑的表格布局
\toprule
~ &GCNII &  APPNP & GPRGNN &  GRAND \\
~ & \multicolumn{4}{c}{\cellcolor[HTML]{E5F2FC}+\agmixup} \\ 
\midrule
\multirow{2}{*}{\textbf{Cora}}  & $83.80_{\pm 1.48}$ & $82.17_{\pm 1.44}$ & $82.12_{\pm  1.11}$ & $81.21_{\pm 1.09}$ \\

~ 
& \cellcolor[HTML]{E5F2FC}$\textbf{84.26}_{\pm 1.10}$ 
& \cellcolor[HTML]{E5F2FC}$\textbf{84.46}_{\pm 0.81}$ 
& \cellcolor[HTML]{E5F2FC}$\textbf{83.95}_{\pm 0.95}$ 
& \cellcolor[HTML]{E5F2FC}$\textbf{83.18}_{\pm 1.17}$ \\

\midrule
 
\multirow{2}{*}{\textbf{Cite.}}  & $70.54_{\pm 0.59}$  & $70.23_{\pm 0.92}$ & $70.03_{\pm 0.52}$ & $69.74_{\pm 0.83}$ \\

~ 
& \cellcolor[HTML]{E5F2FC}$\textbf{72.36}_{\pm 0.74}$ 
& \cellcolor[HTML]{E5F2FC}$\textbf{72.17}_{\pm 0.65}$  
& \cellcolor[HTML]{E5F2FC}$\textbf{71.51}_{\pm 0.85}$  
& \cellcolor[HTML]{E5F2FC}$\textbf{71.10}_{\pm 0.74}$ \\

\midrule

\multirow{2}{*}{\textbf{Pub.}}  & $78.51_{\pm 1.82}$  & $79.08_{\pm 1.65}$ & $79.03_{\pm 1.93}$ & $77.17_{\pm 1.24}$ \\

~ 
& \cellcolor[HTML]{E5F2FC}$\textbf{83.50}_{\pm 1.69}$ 
& \cellcolor[HTML]{E5F2FC}$\textbf{81.18}_{\pm 1.43}$  
& \cellcolor[HTML]{E5F2FC}$\textbf{81.95}_{\pm 1.81}$  
& \cellcolor[HTML]{E5F2FC}$\textbf{81.48}_{\pm 2.48}$ \\

\bottomrule
%\addlinespace
    \multicolumn{5}{l}{\footnotesize{\agmixup's results are listed in the colored rows.}}
    
\end{tabular}
}
\end{table}
In this section, we extend our \agmixup to four advanced GNNs, i.e., GCNII~\cite{gcnii}, APPNP~\cite{appnp}, GPRGNN~\cite{gprgnn}, and GRAND~\cite{grand}, on Cora, Citeseer, and Pubmed datasets. We fix the number of layers of GCNII at 64 and 8 for the other models. The results are summarized in Table~\ref{apptab:advanced_gnns}. The results demonstrate substantial improvements in node classification accuracy with \agmixup across all advanced models when compared to their baseline performances. Notably, on the Cora dataset, \agmixup enhances the GCNII performance from 83.80\% to 84.26\%, and APPNP obtains an increase from 82.17\% to 84.46\%. Similarly, significant gains are also observed in the Citeseer dataset: the baseline accuracy of GPRGNN improves from 70.03\% to 71.51\% with \agmixup. The consistent improvement across various GNNs and datasets reveals the effectiveness and versatility of \agmixup, making it a valuable data augmentation technique for GNNs.

%%%%%%%%%%% Results on Fewer Labels %%%%%%%%%%%
\subsection{Results on Fewer Labels}
\label{appsec:limited}
 \begin{figure*}[!ht]
	 \centering
	 \subfigure[Cora]{
	 			\includegraphics[width=0.95\linewidth]{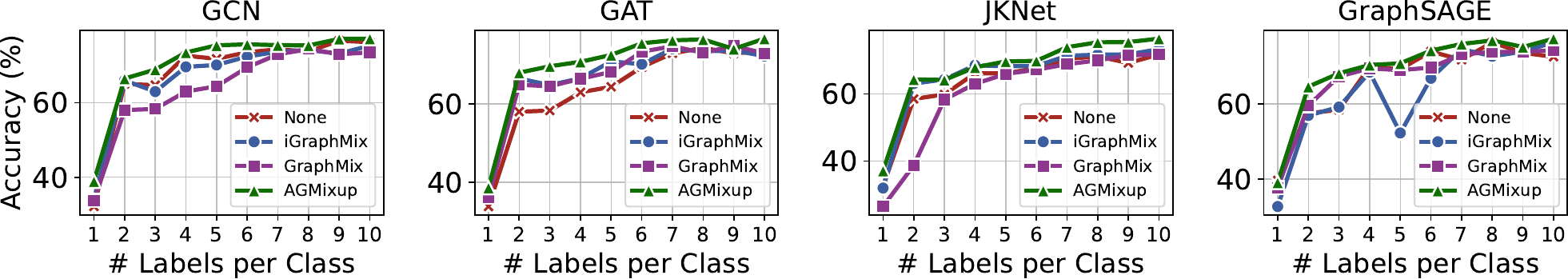}
		}
	\subfigure[Citeseer]{
	 			\includegraphics[width=0.95\linewidth]{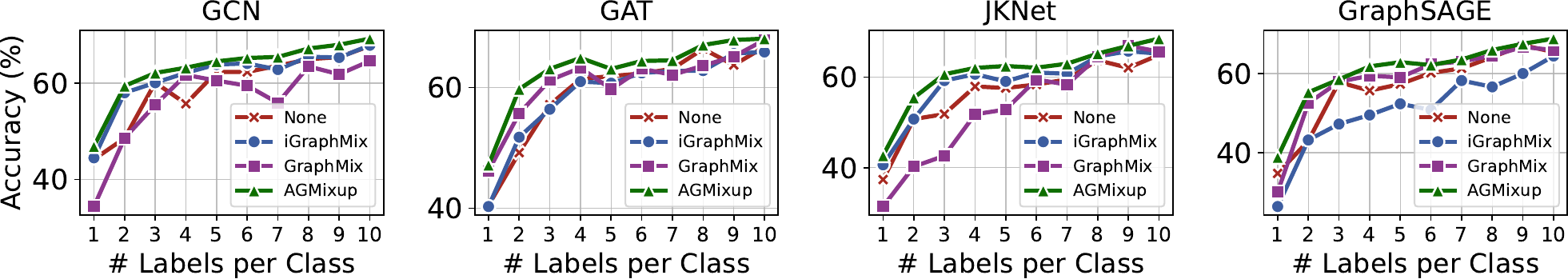}
		}
	\subfigure[Pubmed]{
	 			\includegraphics[width=0.95\linewidth]{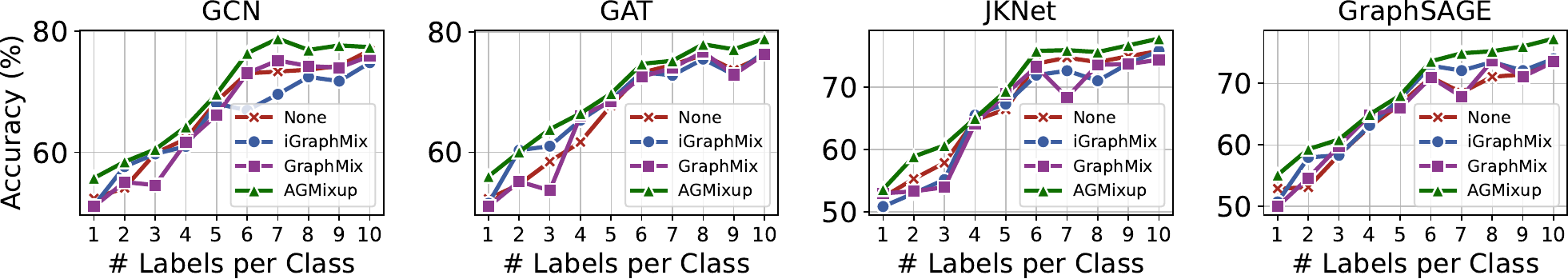}
		}
     \caption{Performance comparison with limited labels using Cora, Citeseer, and Pubmed datasets.}
     \label{appfig:limited}
\end{figure*}

In extending our \agmixup to a limited labels setting, we conduct experiments across various label budgets, ranging from 1 to 10 labels per class. This analysis is visualized in Fig.~\ref{appfig:limited}. We compare our \agmixup against iGraphMix and GraphMix, using GCN, GAT, JKNet, and GraphSAGE as the backbone models, on three datasets: Cora, Citeseer, and Pubmed. The results consistently demonstrate that \agmixup significantly enhances model performance in scenarios with sparse labels. For instance, on the Cora dataset, \agmixup improves GraphSAGE performance from 72.50\% to 77.32\% with ten labels per class, outperforming other methods such as iGraphMix and GraphMix. Similar trends are observed with other models and datasets, where \agmixup not only consistently outperforms the baseline and other mixup methods, but also shows substantial gains as the number of labels increases. This indicates that \agmixup is effective in leveraging limited label information, significantly enhancing GNNs under label scarcity. This is critical for practical applications where acquiring extensive labeled data can be challenging.

%%%%%%%%%%% Results on ogbn-products %%%%%%%%%%%
\subsection{Results on ogbn-products}
\label{appsec:products}

%\begin{wraptable}[14]{r}{0.40\textwidth}
%\vspace{-0.4cm}
 \begin{table}[!t]
\caption{Results on ogbn-products. }
\label{apptab:products}
\centering
\renewcommand\arraystretch{1.3} % 行间距
\setlength\tabcolsep{3pt} % 列间距
\resizebox{0.4\textwidth}{!}{%
\begin{tabular}{@{}lcccc@{}} %@{}:指定空格来消除列之间的间距，从而实现紧凑的表格布局
\toprule
~ & Train &  Valid & Test \\
\midrule
GCN  & 88.30 & 87.58 & 70.48 \\
~+\agmixup 
& $\textbf{89.70}$ & $\textbf{88.87}$ & $\textbf{72.22}$ \\
\cellcolor[HTML]{FDECEE}Imp. (\%)
& \cellcolor[HTML]{FDECEE} 1.58 $\uparrow$
& \cellcolor[HTML]{FDECEE} 1.47 $\uparrow$ 
& \cellcolor[HTML]{FDECEE} 2.46 $\uparrow$ \\

\midrule
 
GraphSAGE  & 89.53  & 88.32 & 71.56 \\

~+\agmixup 
& $\textbf{91.16}$ & $\textbf{89.59}$ & $\textbf{73.66}$ \\

\cellcolor[HTML]{FDECEE}Imp. (\%) 
& \cellcolor[HTML]{FDECEE} 1.82 $\uparrow$ 
& \cellcolor[HTML]{FDECEE} 1.43 $\uparrow$  
& \cellcolor[HTML]{FDECEE} 2.93 $\uparrow$ \\

\bottomrule
    
\end{tabular}
}
\end{table}

In this section, we evaluate our \agmixup on the large-scale graph, i.e., ogbn-products. We fix the learning rate at 0.001, the dropout rate at 0.2, the hidden dimensionality at 128, and the weight decay rate at 0. For the backbone models, i.e, GCN and GraphSAGE, we fix the number of layers at 2. We train each model with or without our \agmixup for 500 epochs. The results are obtained based on the best performance on the validation set. \agmixup on the ogbn-products dataset has demonstrated significant improvements across all metrics of performance, including training, validation, and test accuracies. As indicated in Table~\ref{apptab:products}, the incorporation of \agmixup has enhanced the baseline models' performance substantially. For the GCN model, there is an improvement of 1.58\%, 1.47\%, and 2.46\% in training, validation, and test accuracies, respectively. Similarly, GraphSAGE shows an increase of 1.82\%, 1.43\%, and 2.93\% in the same metrics. The significant improvements noted in both GCN and GraphSAGE models demonstrate \agmixup's suitability for large-scale applications.

\input{tables/comp_aug.tex}

%%%%%%%%%%% Comparison between Other Graph Augmentation Methods %%%%%%%%%%%
\subsection{Comparison against Other Graph Augmentation Methods}
\label{appsec:comp_aug}
In this section, we compare the effectiveness of two established graph augmentation techniques, i.e., DropEdge~\cite{dropedge} and PairNorm~\cite{pairnorm}, with our \agmixup. Our results, summarized in Table~\ref{apptab:comp_aug}, indicate that \agmixup consistently outperforms both DropEdge and PairNorm across all models and datasets in terms of node classification accuracy. For instance, on the Cora dataset using the GCN model, \agmixup achieves an accuracy of 83.61\%, which is significantly higher than 81.06\% and 81.53\% obtained with DropEdge and PairNorm, respectively. This trend of superior performance with \agmixup is consistent across other models and datasets as well. DropEdge randomly removes edges during training, which helps in mitigating overfitting and improving model robustness. PairNorm, on the other hand, stabilizes the training of GNNs by normalizing node features across the graph. While both techniques offer improvements over baseline models, they do not fundamentally introduce new, informative examples during training. \agmixup, by interpolating not only features but also the graph structure between subgraphs, effectively generates novel and informative training examples.

\subsection{Hyperparameter Analysis}
\label{appsec:hyper}

\begin{figure*}[!p]
	 \centering
	 \subfigure[Cora]{
	 			\includegraphics[width=0.95\linewidth]{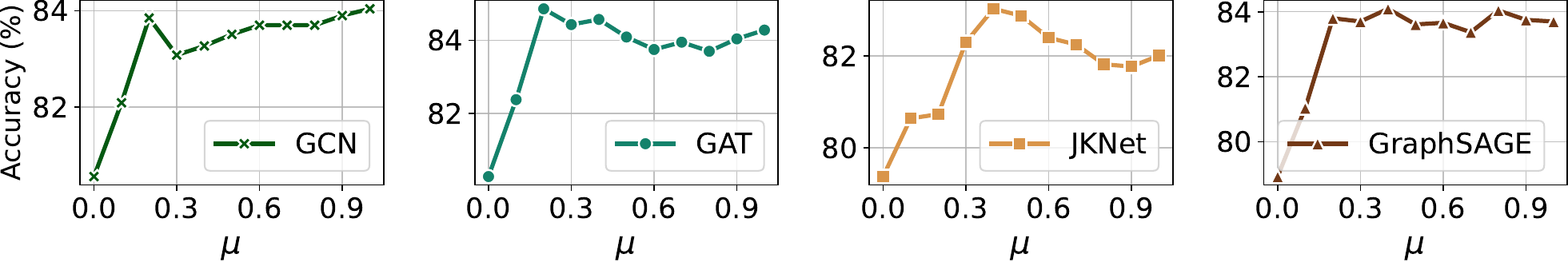}
		}
	\subfigure[Citeseer]{
	 			\includegraphics[width=0.95\linewidth]{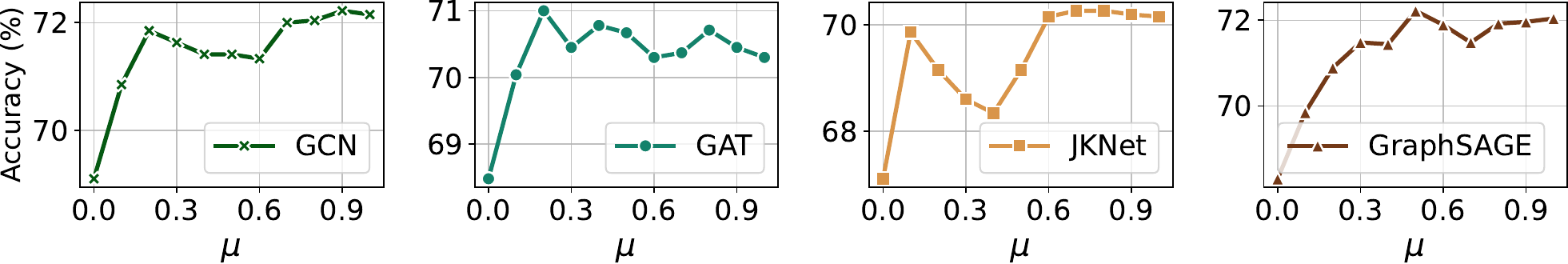}
		}
%	\hfil
	\subfigure[Pubmed]{
	 			\includegraphics[width=0.95\linewidth]{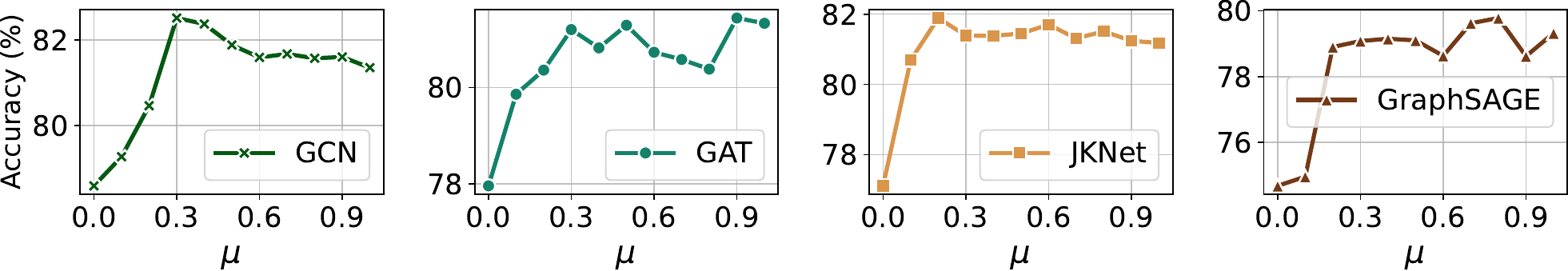}
		}
     \caption{Hyperparameter analysis on $\mu$ using Cora, Citeseer, and Pubmed datasets.}
     \label{appfig:hyper_mu}
\end{figure*}

\paragraph{Weight of Mixup Loss.}
In Fig.~\ref{appfig:hyper_mu}, we conduct a hyperparameter analysis on the balance factor $\mu$, which weighs the significance of the mixup loss, varying from 0 to 1 with an interval of 0.1. We examine its impact on the performance of various GNNs, i.e., GCN, GAT, JKNet, and GraphSAGE, on the Cora and Citeseer datasets. The overall results indicate that increasing $\mu$ generally leads to an improvement in model performance, peaking at intermediate values before stabilizing or slightly declining. Specifically, for the Cora dataset, optimal performance is often observed as $\mu$ is near 0.9, with GCN and GAT showing peak accuracies at 84.04\% and 84.28\%, respectively. Similarly, on the Citeseer dataset, performance enhancements are noted with increasing $\mu$, with GCN and GraphSAGE reaching their best accuracies at 72.22\% and 72.04\%. These trends suggest that a higher weight on mixup loss contributes positively up to a certain threshold, beyond which the benefits plateau or marginally decrease. This indicates the effectiveness of integrating \agmixup as a significant regularizer of the classification loss, enhancing generalization by effectively blending node features and structural information across the graphs.

\begin{figure*}[!htbp]
	 \centering
	 \subfigure[Cora]{
	 			\includegraphics[width=0.95\linewidth]{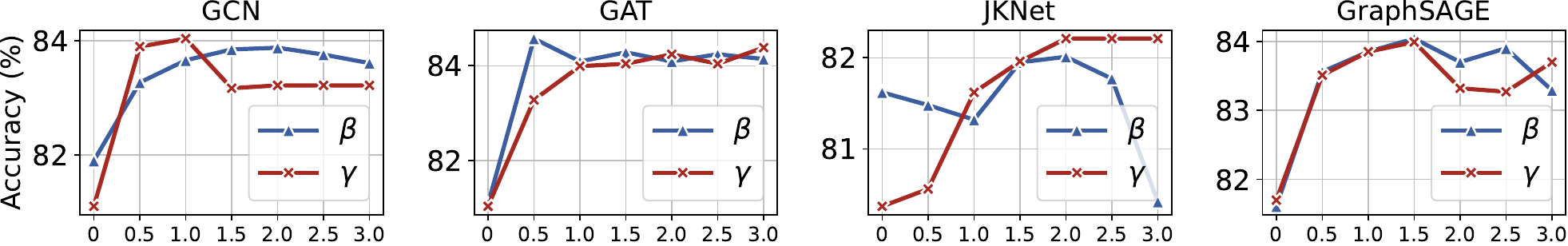}
		}
	\subfigure[Citeseer]{
	 			\includegraphics[width=0.95\linewidth]{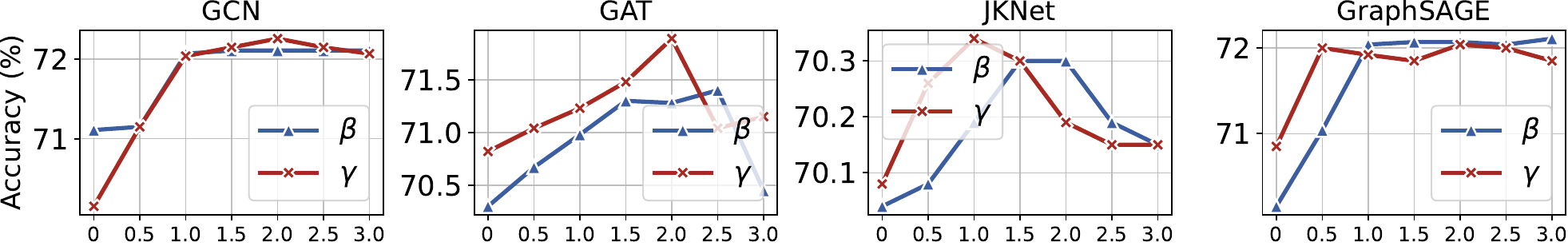}
		}
%	\hfil
	\subfigure[Pubmed]{
	 			\includegraphics[width=0.95\linewidth]{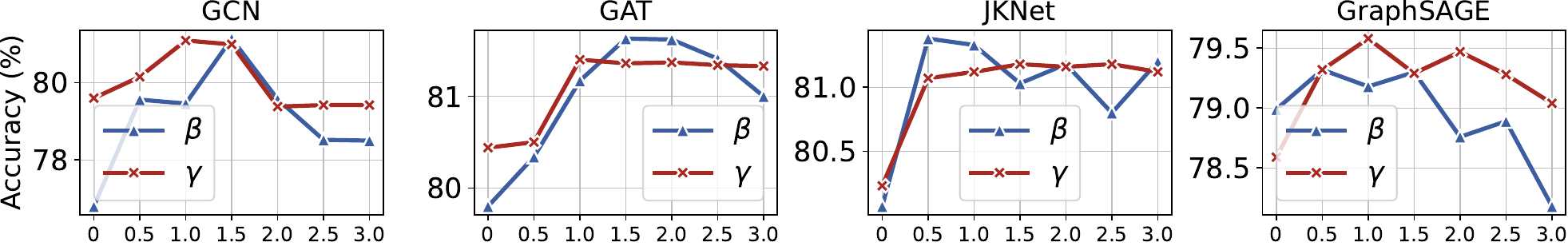}
		}
     \caption{Hyperparameter analysis on $\beta$ and $\gamma$ using Cora, Citeseer, and Pubmed datasets.}
     \label{appfig:hyper_beta_gamma}
\end{figure*}

\paragraph{Sensitivity to Similarity and Uncertainty.}
In Fig.~\ref{appfig:hyper_beta_gamma}, we analyze the effect of two critical hyperparameters, $\gamma$, and $\beta$, on the performance of various GNNs across three datasets: Cora, Citeseer, and Pubmed. $\gamma$, which adjusts the sensitivity to contextual similarity, shows that a moderate increase generally leads to peak performance in accuracy, which then stabilizes or slightly declines as $\gamma$ continues to increase. This trend indicates that while sensitivity to dissimilarity helps in avoiding overly exploratory mixup operations for subgraphs that are not similar, too high a $\gamma$ might restrict the beneficial aspects of mixup by being overly conservative. $\beta$, which modulates sensitivity to the uncertainty difference between subgraphs, often shows an improvement in performance up to a certain point. The results suggest that while initially, the increasing $\beta$ is beneficial to addressing the uncertainty effectively by adapting the mixup operation, excessively high values might lead to diminishing returns, possibly due to overemphasis on uncertain and less informative regions.

\subsection{Efficacy Analysis}
\label{appsec:efficacy}
 \begin{figure*}[!htbp]
	 \centering
	 \subfigure[Cora]{
	 			\includegraphics[width=0.99\linewidth]{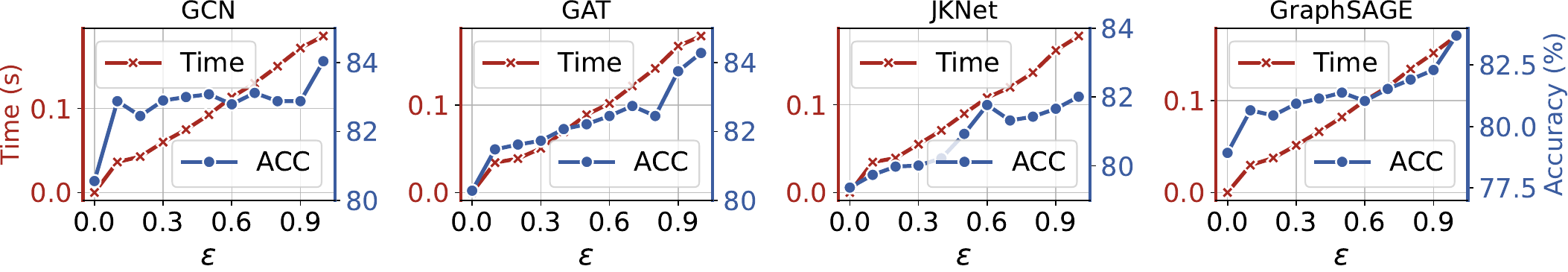}
		}
	\subfigure[Citeseer]{
	 			\includegraphics[width=0.99\linewidth]{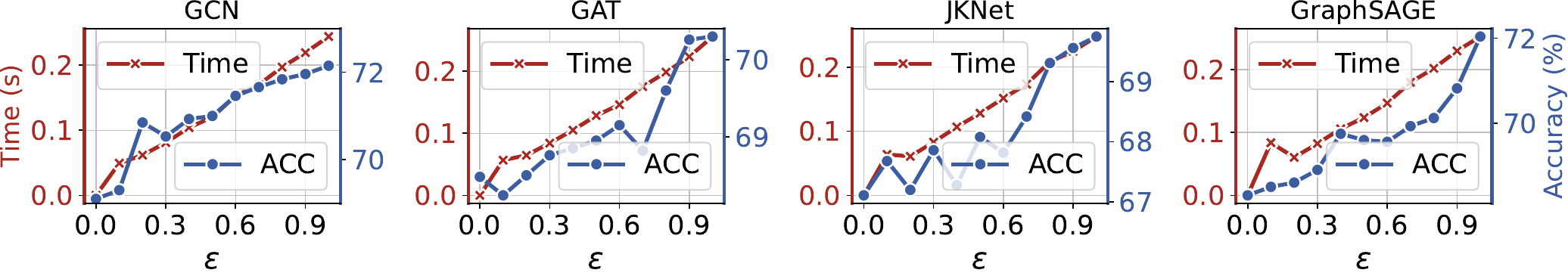}
		}
%	\hfil
	\subfigure[Pubmed]{
	 			\includegraphics[width=0.99\linewidth]{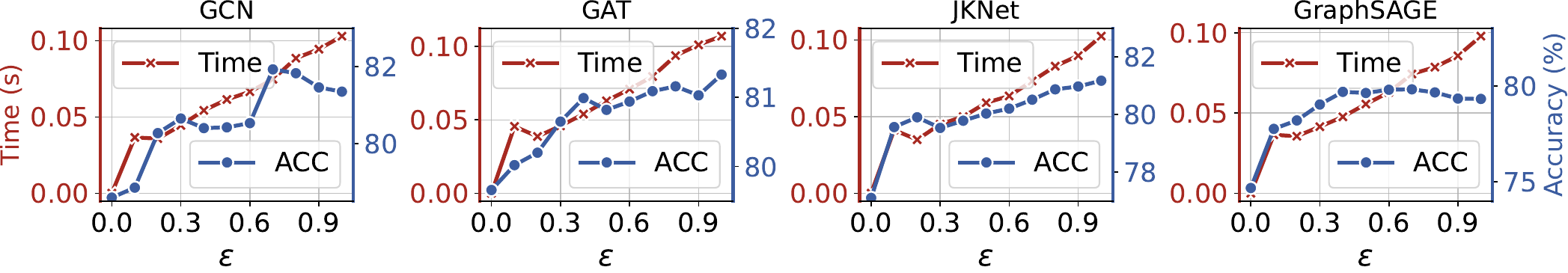}
		}
     \caption{Efficacy analysis using Cora, Citeseer, and Pubmed datasets.}
     \label{appfig:efficacy}
\end{figure*}

In Fig.~\ref{appfig:efficacy}, we analyze the relationship between the time cost per epoch of mixup training and accuracy of \agmixup w.r.t different values of $\epsilon$, ranging from 0 to 1 with an interval of 0.1. This parameter controls the proportion of the dataset involved in the mixup process, directly influencing both the time required for computation and the resulting model accuracy.

Our results demonstrate a clear trend in computational time across various models and datasets as $\epsilon$ increases from 0 to 1. This trend shows the additional computational overhead incurred with the increase in the data proportion used for mixup. Such behavior underscores the expected increase in computational cost as more subgraphs are used for mixup training. Conversely, performance results across models exhibit an improvement with an increase in $\epsilon$, peaking at certain values before plateauing or slightly declining. For instance, on the Pubmed dataset, while GAT reaches its highest accuracy at $\epsilon = 1$, other models like GCN and GraphSAGE often exhibit optimal performance at slightly lower $\epsilon$ values. This indicates that while increasing $\epsilon$ generally benefits model generalization, there exists an optimal range beyond which the performance benefits do not justify the additional computational costs.

These findings suggest a critical trade-off in the application of \agmixup. A lower $\epsilon$ might not fully capitalize on the potential of mixup for enhancing model robustness, whereas a higher $\epsilon$ might lead to unnecessary computational expenditure with minimal gains in performance. Based on our analysis, an $\epsilon$ value in the range of 0.5 to 0.7 typically offers the most balanced approach, enhancing accuracy effectively without imposing excessive computational demands.
\end{document}

%% file: algorithm/alg.tex
\begin{algorithm}[!thp]
\caption{\agmixup's Training Algorithm}
\label{alg:agmixup}

\textbf{Input:} Original graph $\mathcal{G} = \{\mathcal{V}, \mathcal{E}, X\}$, gnn model $g_\theta$ \\
\textbf{Parameters}: Similarity sensitivity ratio $\gamma$, uncertainty sensitivity ratio $\beta$, subgraph size $r$, number of epochs $T$
%\begin{multicols}{2}

\begin{algorithmic}[1]
\STATE Initialize model parameters $\theta$
\FOR{$t = 1$ to $T$}
	\STATE \textcolor{gray}{\# Step 1: Mixup Pair Sampling}
    \STATE Sample mixup pairs $\mathcal{P} = \{(v_{i}, v_{j})_{m}\}^{|\mathcal{V}_{L}|}_{m=1}$ from $\mathcal{V}_{L}$ 
        
    \STATE \textcolor{gray}{\# Step 2: Subgraphs Extracting}
    \FOR{node $v_{i}  \in \mathcal{V}_{L}$}
        \STATE Extract subgraphs $\mathcal{G}^{(r)}_{i}$ of size $r$ around $v_i$ 
    \ENDFOR
 	
 	\STATE \textcolor{gray}{\# Step 3: Mixed Graphs Generating}
    \FOR{each pair of subgraphs $(\mathcal{G}^{(r)}_i, \mathcal{G}^{(r)}_j)$}
    	\STATE \textcolor{gray}{\# Step 3.1: Contextual Similarity-aware Initialization}
        \STATE Calculate $\lambda_{ij}$ via Eq.~(3)
        \STATE \textcolor{gray}{\# Step 3.2: Uncertainty-aware Adjustment}
        \STATE Adjust $\lambda_{ij}$ via Eq.~(4)
        \STATE \textcolor{gray}{\# Step 3.3: Mixed Graph Generation} 
        \STATE Generate the mixed graph $\mixpoint{\mathcal{G}}{ij}$ via Eq.~(2)   
    \ENDFOR
    
    \STATE \textcolor{gray}{\# Step 4: Forward Pass (Original Graph)}
	\STATE Compute prediction $\hat{\mathcal{Y}} = g_\theta(\mathcal{G})$
        
    \STATE \textcolor{gray}{\# Step 5: Forward Pass (Mixed Graphs)}
    \FOR{each mixed subgraph $\mathcal{G}_{ij}$}
        \STATE Compute prediction $\hat{\mathcal{Y}}_{ab} = g_\theta(\mathcal{G}_{ij})$
    \ENDFOR
    
    \STATE \textcolor{gray}{\# Step 6: Compute Loss}
    \STATE Compute loss via Eq.~(7)
    
    \STATE \textcolor{gray}{\# Step 7: Backward Pass and Update}
    \STATE Compute gradients $\nabla_\theta \mathcal{L}$
    \STATE Update parameters $\theta \leftarrow \theta - \eta \nabla_\theta \mathcal{L}$
\ENDFOR
\STATE \textbf{Return} Trained model $g_\theta$
\end{algorithmic}

%\end{multicols}
\end{algorithm}

%% file: tables/comp_aug.tex
\begin{table*}[ht!]
    \centering
    \caption{Comparison between DropEdge, PairNorm, and \agmixup.}
%    \resizebox{1\linewidth}{!}{
    \begin{tabular}{l c c c c c}
    \hline
    \toprule
    
    ~ & Cora & Citeseer & Pubmed & CS & Physics  \\

	\midrule
	GCN  
    & 80.51$_{\pm1.39}$ 
    & 69.53$_{\pm0.96}$ 
    & 78.23$_{\pm1.64}$ 
    & 91.22$_{\pm0.83}$ 
    & 93.51$_{\pm0.34}$ \\
    
    ~ + DropEdge 
    & 81.06$_{\pm0.98}$ 
    & 70.30$_{\pm0.84}$ 
    & 78.94$_{\pm1.08}$ 
    & 91.44$_{\pm0.62}$ 
    & 93.92$_{\pm0.86}$ \\
    
    ~ + PairNorm 
    & 81.53$_{\pm1.93}$ 
    & 69.68$_{\pm1.21}$ 
    & 78.95$_{\pm1.92}$ 
    & 91.46$_{\pm1.62}$ 
    & 93.74$_{\pm0.77}$ \\
    
    ~ + \textbf{\agmixup (\textit{ours})}
    & \textbf{83.61}$_{\pm0.39}$ & \textbf{70.63}$_{\pm1.54}$ & \textbf{81.90}$_{\pm0.40}$ & \textbf{92.35}$_{\pm0.92}$ & \textbf{94.45}$_{\pm0.42}$ \\ 
    
%    \midrule
	\cmidrule(r){1-1} \cmidrule(r){2-4}  \cmidrule(r){5-6} 
	    
    GAT 
    & 80.02$_{\pm1.15}$ 
    & 68.38$_{\pm1.35}$ 
    & 77.28$_{\pm1.80}$ 
    & 90.06$_{\pm0.80}$ 
    & 92.59$_{\pm0.64}$ \\
    
    ~ + DropEdge 
    & 81.61$_{\pm0.92}$ 
    & 68.96$_{\pm1.04}$ 
    & 77.41$_{\pm1.55}$ 
    & 90.26$_{\pm0.84}$ 
    & 92.21$_{\pm0.98}$ \\
    
    ~ + PairNorm 
    & 80.97$_{\pm1.25}$ 
    & 68.06$_{\pm0.94}$ 
    & 77.12$_{\pm1.71}$ 
    & 90.52$_{\pm1.15}$ 
    & 92.64$_{\pm1.23}$ \\
    
    ~ + \textbf{\agmixup (\textit{ours})}
    & \textbf{83.14}$_{\pm0.54}$ & \textbf{69.72}$_{\pm1.51}$ & \textbf{81.60}$_{\pm0.51}$ & \textbf{91.92}$_{\pm1.41}$ & \textbf{94.31}$_{\pm0.63}$ \\ 
    
%    \midrule
	\cmidrule(r){1-1} \cmidrule(r){2-4}  \cmidrule(r){5-6}  
		    
    GraphSAGE 
    & 79.28$_{\pm1.20}$ 
    & 68.36$_{\pm0.89}$ 
    & 75.72$_{\pm1.84}$ 
    & 91.95$_{\pm0.36}$ 
    & 93.18$_{\pm0.47}$ \\
    
    ~ + DropEdge 
    & 80.65$_{\pm1.02}$ 
    & 69.25$_{\pm0.72}$ 
    & 75.93$_{\pm0.99}$ 
    & 92.26$_{\pm0.41}$ 
    & 92.97$_{\pm0.63}$ \\
    
    ~ + PairNorm 
    & 78.64$_{\pm1.42}$ 
    & 67.20$_{\pm1.57}$ 
    & 75.04$_{\pm1.96}$ 
    & 90.86$_{\pm0.81}$ 
    & 92.27$_{\pm0.98}$ \\
    
    ~ + \textbf{\agmixup (\textit{ours})} 
    & \textbf{82.67}$_{\pm1.16}$ & \textbf{70.04}$_{\pm1.08}$ & \textbf{79.84}$_{\pm0.74}$  & \textbf{92.31}$_{\pm0.50}$ & \textbf{94.28}$_{\pm0.64}$\\ 
    
    \bottomrule
	    
	\label{apptab:comp_aug}
    \end{tabular}
%    }
    
\end{table*}